\documentclass[10pt,twocolumn,letterpaper]{article}

\usepackage[pagenumbers]{cvpr} 

\usepackage[accsupp]{axessibility}
\usepackage{graphicx}
\usepackage{amsmath}
\usepackage{amssymb}
\usepackage{booktabs}

\usepackage[noend]{algpseudocode}
\usepackage{algorithmicx,algorithm}

\usepackage{threeparttable}
\usepackage{multirow,booktabs}
\usepackage{adjustbox}

\usepackage[breaklinks,colorlinks]{hyperref}

\usepackage[capitalize]{cleveref}
\crefname{section}{Sec.}{Secs.}
\Crefname{section}{Section}{Sections}
\Crefname{table}{Table}{Tables}
\crefname{table}{Tab.}{Tabs.}

\usepackage{color}
\definecolor{myGreen}{RGB}{10,180,10}

\usepackage{listings}
\usepackage[dvipsnames]{xcolor}

\definecolor{codegreen}{rgb}{0,0.6,0}
\definecolor{codegray}{rgb}{0.5,0.5,0.5}
\definecolor{codepurple}{rgb}{0.58,0,0.82}
\definecolor{backcolour}{rgb}{0.95,0.95,0.92}

\lstdefinestyle{mystyle}{
    backgroundcolor=\color{backcolour},   
    commentstyle=\color{codegreen},
    keywordstyle=\color{magenta},
    numberstyle=\tiny\color{codegray},
    stringstyle=\color{codepurple},
    basicstyle=\ttfamily\footnotesize,
    breakatwhitespace=false,         
    breaklines=true,                 
    captionpos=b,                    
    keepspaces=true,                 
    numbers=left,                    
    numbersep=5pt,                  
    showspaces=false,                
    showstringspaces=false,
    showtabs=false,                  
    tabsize=2
}

\begin{document}

\title{SimAN: Exploring Self-Supervised Representation Learning of Scene Text \\
via Similarity-Aware Normalization}

\author{Canjie Luo\textsuperscript{\rm 1}, Lianwen Jin\textsuperscript{\rm 1,2,}\thanks{Corresponding author.}{ \ },  Jingdong Chen\textsuperscript{\rm 3}\\
\textsuperscript{\rm 1}South China University of Technology, 
\textsuperscript{\rm 2}Peng Cheng Laboratory, 
\textsuperscript{\rm 3}Ant Group\\
{\tt\small \{canjie.luo, lianwen.jin\}@gmail.com, jingdongchen.cjd@antgroup.com}}
\maketitle

\begin{abstract}
Recently self-supervised representation learning has drawn considerable attention from the scene text recognition community. Different from previous studies using contrastive learning, we tackle the issue from an alternative perspective, i.e., by formulating the representation learning scheme in a generative manner. Typically, the neighboring image patches among one text line tend to have similar styles, including the strokes, textures, colors, etc. Motivated by this common sense, we augment one image patch and use its neighboring patch as guidance to recover itself. Specifically, we propose a \underline{Sim}ilarity-\underline{A}ware \underline{N}ormalization (SimAN) module to identify the different patterns and align the corresponding styles from the guiding patch. In this way, the network gains representation capability for distinguishing complex patterns such as messy strokes and cluttered backgrounds. Experiments show that the proposed SimAN significantly improves the representation quality and achieves promising performance. Moreover, we surprisingly find that our self-supervised generative network has impressive potential for data synthesis, text image editing, and font interpolation, which suggests that the proposed SimAN has a wide range of practical applications.

\end{abstract}

\section{Introduction}
The computer vision community has witnessed the great success of supervised learning over the last decade. However, the supervised learning methods heavily rely on labor-intensive and expensive annotations. Otherwise, they might suffer from generalization problems. Recently self-supervised representation learning has become a promising alternative and is thus attracting growing interest~\cite{Jing2020Self,liu2021self}. It has been shown that the self-supervised representations can benefit subsequent supervised tasks~\cite{he2020momentum,chen2021empirical,chen2020simple,chen2020big,chen2021exploring,chen2020generative}. 

Despite the fast-paced improvements of representation learning on single object recognition/classification tasks, the field of scene text recognition is meeting extra challenges. For instance, multiple characters in one image cannot be regarded as one entity~\cite{zhang2019sequence,luo2020learn}. Directly adopting current non-sequential contrastive learning schemes for sequence-like characters~\cite{shi2017end} usually leads to performance deterioration~\cite{aberdam2021sequence}. This suggests the gap between the non-sequential and sequential schemes. Therefore, it is desirable to design a specific representation learning scheme for scene text recognition.

\begin{figure}[t]
  \centering
  \includegraphics[width=0.88\columnwidth]{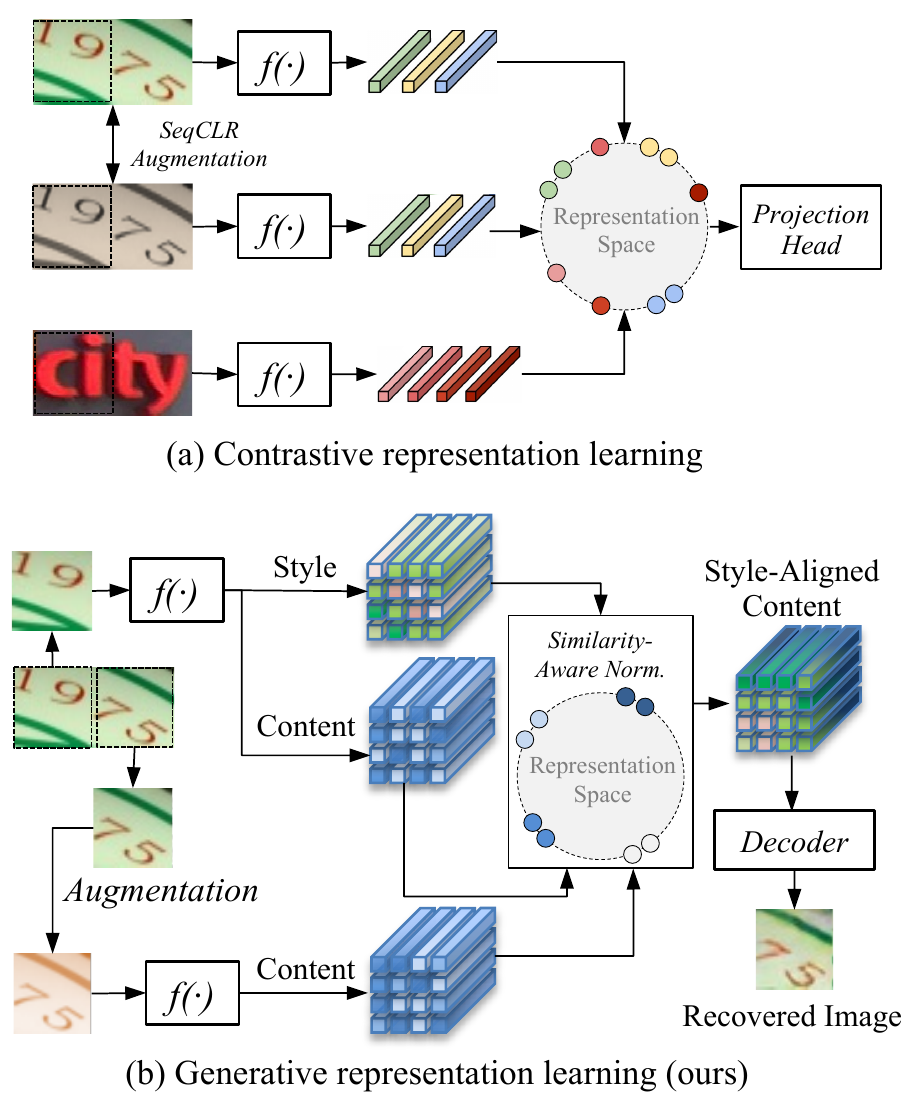}
  \caption{Scene text representation learning in (a) the contrastive and (b) the generative manner (ours). We estimate the similarity of the content representations between the augmented patch and its neighboring patch, and align the corresponding styles to reconstruct the augmented patch. Only high-quality representations are distinguishable so that a precise reconstruction can be achieved.}
  \label{pic-contrastive-generative}
  \vspace{-0.em}
\end{figure}

As a scene text image containing dense characters is significantly different from a natural image, SeqCLR~\cite{aberdam2021sequence} divided one text line into several instances using certain strategies and performed contrastive learning on these instances. The learning scheme is shown in Figure~\ref{pic-contrastive-generative} (a). The SeqCLR designed for sequence-to-sequence visual recognition outperformed the representative non-sequential method SimCLR~\cite{chen2020simple}. Although it brought a huge leap forward, the representation learning of scene text remains a challenging open research problem, where the nature of scene text has not been fully explored. 

Thus, we review several properties of scene text that differ from those of general objects (\textit{e.g.}, face, car, and dog). For instance, one feature that highlights scene text is its constant stroke width~\cite{epshtein2010detecting}. Simultaneously, it is observed that color similarity typically occurs across one text line. These specialties provided cues for hand-crafted features, such as connected components~\cite{neumann2012real}, stroke width transform~\cite{epshtein2010detecting,yao2012detecting}, and maximally stable extremal region trees~\cite{huang2014robust}, which were popular before the dramatic success of deep neural networks. 

In this paper, we explore the representation learning from a new perspective by considering the above unique properties of scene text. The learning scheme is shown in Figure~\ref{pic-contrastive-generative} (b). Specifically, we randomly crop two neighboring image patches from one text line. One patch is augmented and the other one guides the recovery of the augmented one. As one text line usually exhibits consistent styles, including the strokes, textures, colors, \textit{etc.}, the original styles of the augmented patch can be found on the neighboring patch according to similar content patterns. Thus, we propose a \underline{Sim}ilarity-\underline{A}ware \underline{N}ormalization (SimAN) module to align corresponding styles from the neighboring patch by estimating the similarity of the content representations between these two patches. This means that the representations are required to be sufficiently distinguishable so that different patterns can be identified and the corresponding styles can be correctly aligned. Only in this way, the network can produce a precise recovered image patch. Therefore, the proposed SimAN enables high-quality self-supervised representation learning in a generative way. Moreover, we find that our self-supervised network has competitive performance with state-of-the-art scene text synthesis methods~\cite{jaderberg2016reading,gupta2016synthetic,zhan2018verisimilar,long2020unrealtext}. It is also promising to apply SimAN to other visual effect tasks, such as text image editing and font interpolation.

To summarize, our contributions are as follows:
\vspace{-0.5em}
\begin{itemize}
\setlength{\itemsep}{0pt}
\setlength{\parsep}{0.5pt}
\setlength{\parskip}{0.5pt}
\item We propose a generative (opposite of contrastive~\cite{liu2021self}) representation learning scheme by utilizing the unique properties of scene text, which might inspire rethinking the learning of better representations for sequential data like text images. To the best of our knowledge, this is the first attempt for scene text recognition. 
\item We propose a SimAN module, which estimates the similarity of the representations between the augmented image patch and its neighboring patch to align corresponding styles. Only if the representations are sufficiently distinguishable, different patterns can be identified and be aligned with correct styles. Otherwise, the network might result in a wrong recovered image, \textit{e.g.}, in different colors.
\item The proposed SimAN achieves promising representation performance. Moreover, the self-supervised network shows impressive capabilities to synthesize data, edit text images and interpolate fonts, suggesting the broad practical applications of the proposed approach.
\end{itemize}

\section{Related Work}
\subsection{Data Hunger of Scene Text Recognition}
Scene text recognition is a crucial research topic in the computer vision community, because the text in images provides considerable semantic information for us. One important open issue in this field is data hunger. Typically, mainstream scene text recognizers~\cite{shi2018aster,wang2020decoupled,fang2021read} require a large number of annotated data. However, data collection and annotation cost a lot of resources. For instance, annotating a text string is tougher than selecting one option as the ground truth for single object classification datasets, whereas tens of millions of training data are required to gain robustness. Although synthetic data are available, previous studies~\cite{li2019show,zhang2019sequence,kang2020unsupervised,luo2021separating} suggested that there is a gap between real and synthetic data. To mitigate this problem, Zhang \textit{et al.}~\cite{zhang2019sequence} and Kang \textit{et al.}~\cite{kang2020unsupervised} proposed domain adaptation models to utilize unlabeled real data. Our study explores representation learning in a generative way, which is an alternative solution to make use of unlabeled real data.

\subsection{Visual Representation Learning}
In the big data era, tremendous amounts of unlabeled data are available. Making the best use of unlabeled data becomes a crucial topic. Self-supervised representation learning has drawn massive attention owing to its excellent capability of pre-trained feature extraction~\cite{Jing2020Self,liu2021self}. For instance, an encoder trained after a pretext task can extract transferrable features to benefit downstream tasks. We summarize popular methods into two main categories according to their objectives as follows.

\begin{figure*}[t]
  \centering
  \includegraphics[width=1.88\columnwidth]{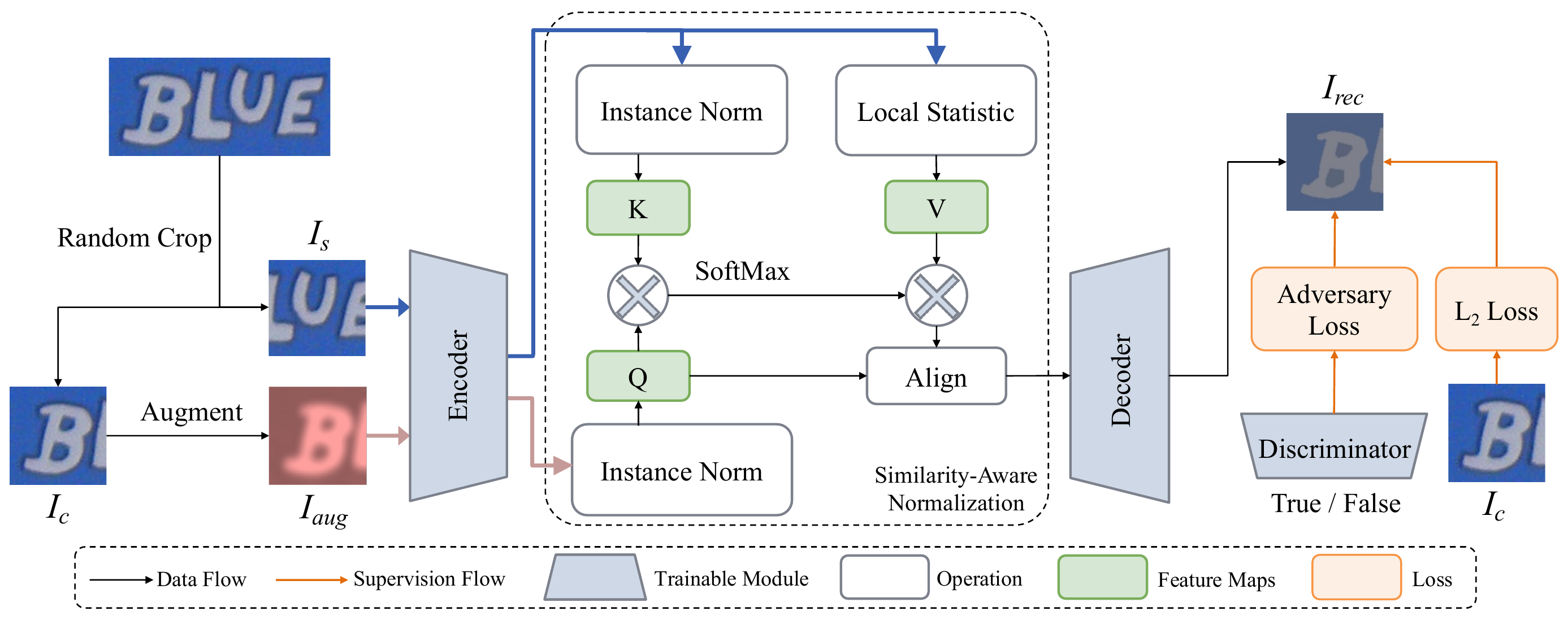}
  \caption{Overview of the proposed generative representation learning scheme. We decouple content and style as two different inputs and guide the network to recover the augmented image. The proposed SimAN module learns to align corresponding styles for different patterns according to the distinguishable representations.}
  \label{pic-overall}
  \vspace{-0.0em}
\end{figure*}

The \textbf{contrastive learning scheme} defines the pretext task as a classification task or a distance measuring task. For instance, the pretext task is to predict relative rotation~\cite{komodakis2018unsupervised} and position~\cite{wei2019iterative}. Recently the similarity measuring pretext task has become dominant, which aims to minimize the distance between the positive pairs while maximizing their distance to the negative ones using a discriminative head~\cite{he2020momentum,chen2020big,chen2020simple,caron2020unsupervised,chen2021empirical}. It is closely related to metric learning. Furthermore, the similarity measuring task using only positive pairs and discarding negative samples~\cite{grill2020bootstrap,chen2021exploring} is also an emerging topic. 

For the field of scene text,  Baek \textit{et al.}~\cite{baek2021if} introduced existing self-supervised techniques ~\cite{komodakis2018unsupervised,he2020momentum} to use unlabeled data but resulted in approximately the same performance. Aberdam \textit{et al.}~\cite{aberdam2021sequence} proposed a contrastive representation learning scheme, termed SeqCLR, to satisfy the sequence-to-sequence structure of scene text recognition. This is the first step towards scene text representation.

The \textbf{generative learning scheme} has not been intensively studied in computer vision. One reason for this may be that the raw image signal is in a continuous and high-dimensional space, unlike the natural language sentences in a discrete space (\textit{e.g.}, words or phrases)~\cite{he2020momentum}. Therefore, it is difficult to define an instance. Although it is possible to model the image pixel by pixel~\cite{van2016pixel}, this theoretically requires much more high-performance clusters~\cite{chen2020generative}. Another solution is the denoising auto-encoder~\cite{vincent2008extracting,van2017neural}, which learns features by reconstructing the (corrupted) input image. 

Our approach falls into the second category of visual representation learning, \textit{i.e.}, the generative learning scheme. We propose a novel representation learning scheme by studying the unique properties of scene text and using an image reconstruction pretext task.

\section{Methodology}
In this section, we first introduce the design of the pretext task and the construction of the training samples. Then, we detail the proposed SimAN module. Finally, we present the objectives of the task and the complete learning scheme. The overall framework is shown in Figure~\ref{pic-overall}. 

\subsection{Training Sample Construction}
Constructing appropriate training samples is critical to the success of the pretext task. We enable the scene text representation learning by recovering an augmented image patch using its neighboring patch as guidance. This design considers the unique properties of scene text, \textit{i.e.}, the styles (\textit{e.g.}, stroke width, textures, and colors) within one text line tend to be consistent. 

The pretext task requires decoupled style and content inputs. As shown in Figure~\ref{pic-overall}, given an unlabeled text image $I \in \mathbb{R} ^ {3 \times H \times W}$(the width $W$ is required to be larger than two times of height $H$), we randomly crop two neighboring image patches $I_s, I_c \in \mathbb{R} ^ {3 \times H \times H}$ as style and content input, respectively. This ensures sufficient differences in content between the two patches. Even if the neighboring patches might contain a same characters, their positions are different. Then, we augment (blurring, random noise, color changes, \textit{etc.}) the content patch $I_c$ as $I_{aug}$ to make its style different from the style patch $I_s$. Finally, the pretext task takes $I_{aug}$ as content input and $I_s$ as the style guidance to recover an image $I_{rec}$. The source content patch $I_c$ serves as supervision.

\textbf{Discussion} As our pretext task is recovering an augmented patch under the guidance of its neighboring patch, the visual cues should be consistent in both patches. Some spatial augmentation strategies, such as elastic transformation, might break the consistency and lead to failed training. For instance, it might bring changes to the stroke width. The excessively distorted strokes are also diverse from the source font style. Therefore, we avoid all of the spatial transformation augmentation methods that are widely used for self-supervised representation learning. This is also a significant difference with previous study SeqCLR~\cite{aberdam2021sequence}.

\subsection{Similarity-Aware Normalization}
Previous studies~\cite{huang2017arbitrary,karras2019style} revealed that the statistics of feature maps, including mean and variance, can represent styles. Based on this finding, we perform instance normalization (IN)~\cite{Dmitry2016Instance,huang2017arbitrary} on the feature maps to remove the style and obtain content representations as key ($K$, from $I_s$) and query ($Q$, from $I_{aug}$) as
\begin{equation}
\small
K = \operatorname{IN}\big(\operatorname{Encoder}(I_s)\big), Q = \operatorname{IN}\big(\operatorname{Encoder}(I_{aug})\big),
\end{equation}
where the $K$ and $Q$ are normalized feature maps with spatial scale $\mathbb{R} ^ {C_F \times H_F \times W_F}$. The $\operatorname{IN}(\cdot)$ is compute as
\begin{equation}
\small
\operatorname{IN}(x) = \frac{x - \mu(x)}{\sqrt{\sigma(x)^{2} + \epsilon}},
\end{equation}
where $\mu(\cdot)$ and $\sigma(\cdot)$ respectively compute the mean and standard deviation, performing independently for each channel and each sample. 

For the local style representations, we extract eight-neighborhood mean and standard deviation at position $(i, j)$ on the $c$-th channel of the feature maps as
\begin{equation}
\small
\mu_{c, i, j}=\frac{1}{9} \sum\nolimits_{p, q \in \mathcal{N}_{i, j}} x_{c, p, q},
\end{equation}
\begin{equation}
\small
\sigma_{c, i, j}=\frac{1}{3}\sqrt{ \sum\nolimits_{p, q \in \mathcal{N}_{i, j}}\left(x_{c, p, q}-\mu_{c, i, j}\right)^{2}},
\end{equation}
where $\mathcal{N}_{i, j}$ is the position set comprising of the eight-neighborhood around the position $(i, j)$ and itself. Here $\mu, \sigma \in \mathbb{R} ^ {C_F \times H_F \times W_F}$ serve as value ($V$, from $I_s$).

Then the statistics $\mu$ and $\sigma$ is adaptively rearranged according to the similarity between the patterns of the two inputs by (here $K$, $Q$, $\mu$ and $\sigma$ are reshaped to $\mathbb{R} ^ {C_F \times H_F W_F}$)
\begin{equation}
\small
\mu^{\prime}=\mu \operatorname{Softmax}\left(\frac{K^{\mathrm{T}}Q}{\sqrt{d_{k}}}\right),\sigma^{\prime}=\sigma \operatorname{Softmax}\left(\frac{K^{\mathrm{T}}Q}{\sqrt{d_{k}}}\right),
\end{equation}
where $d_{k}$ is the dimension of the input $K$. The $\mu^{\prime}$ and $\sigma^{\prime}$ are reshaped to $\mathbb{R} ^ {C_F \times H_F \times W_F}$.

Finally, we perform a reverse process of $\operatorname{IN}(\cdot)$ to align rearranged styles to each position for image recovery as 
\begin{equation}
\small
Q_{c, i, j}^{\prime} = Q_{c, i, j}\sigma_{c, i, j}^{\prime} + \mu_{c, i, j}^{\prime},
\end{equation}
\begin{equation}
\small
I_{rec} = \operatorname{Decoder}(Q^{\prime}).
\end{equation}

As the proposed SimAN integrates styles and contents to recover an image, it enables representation learning. If the encoder produces meaningless content or  style representations, the decoder cannot correctly recover the source image. For instance, the unidentifiable content representations will confuse the style alignment and result in a messy image. The inaccurate style representations will lead to color distortions. In a word, the image reconstruction objective requires effective representations of both content and style. 

\subsection{Learning Scheme}
As we formulate the pretext task as image reconstruction, the source patch $I_c$ can serves as supervision. We minimize the distance between the recovered image $I_{rec}$ and target image $I_c$ as
\begin{equation}
\small
\mathcal{L}_2 = \| I_{rec} - I_c \|_{2}^{2}.
\end{equation}

Simultaneously, we adopt a widely used adversarial objective to minimize the distribution shift between the generated and real data:
\begin{equation}
\small
\min _{D} \mathcal{L}_{adv}=\mathbb{E} \big [ \big (D (I_s)-1 \big)^{2} ]+\mathbb{E} [ \big (D (I_{rec} ) \big )^{2} \big],
\end{equation}
\begin{equation}
\small
\min _{\text {Encoder, Decoder }} \mathcal{L}_{adv}=\mathbb{E}\big [ (D (I_{rec})-1\big)^{2}\big],
\end{equation}
where $D$ denotes a discriminator.

The complete learning scheme is shown in Algorithm~\ref{alg-scheme}. The encoder/decoder and discriminator are alternately optimized to achieve adversarial training.

\begin{algorithm}[h]
\caption{Representation Learning Scheme}
\label{alg-scheme}
\hspace*{0.02in} {\bf Input:} $\operatorname{Encoder}$, $\operatorname{Decoder}$, Discriminator $D$\\
\hspace*{0.02in} {\bf Output:} $\operatorname{Encoder}$, $\operatorname{Decoder}$
\begin{algorithmic}[1]
\For{iteration t = 0, 1, 2, ..., T} 
	\State Sample a mini-batch $\{I_i\}_{i=1}^{B}$ from unlabeled data
	\For{each $I_i$}
	\State Randomly crop $I_s$ and $I_c$, augment $I_c$ as $I_{aug}$
	\EndFor
	\State Forward $\operatorname{Encoder}$, $\operatorname{SimAN}$ and $\operatorname{Decoder}$
	\State Compute loss for $\{I_{rec,i}\}_{i =1}^{B}$ 
	\State Update $D$ using $\min \limits_{D} \mathcal{L}_{adv}$
	\State Update $\operatorname{Encoder}$ and $\operatorname{Decoder}$ using 
	\begin{center} $\min \limits_{\text {Encoder, Decoder }} \mathcal{L}_{adv} + \lambda \mathcal{L}_2$ \end{center}
	\State (The $\lambda$ is empirically set to 10.)
\EndFor
\end{algorithmic}
\end{algorithm}

\section{Experiments}
In this section, we conduct extensive experiments to validate the effectiveness of the proposed approach. First, we compare the quality of the learned representations with that of the previous study SeqCLR~\cite{aberdam2021sequence}. Then, we study the performance of our approach by using a semi-supervised setting, where we pre-train the encoder using unlabeled data and fine-tune it using partially labeled data. Finally, we show the potential of our generative approach for other visual tasks. For instance, we attempt to synthesize diverse data to train a robust recognizer. Moreover, we compare our self-supervised model with mainstream supervised models on the text image editing task. We also demonstrate some promising visual effects on font interpolation.

\begin{figure*}[t]
  \centering
  \includegraphics[width=1.95\columnwidth]{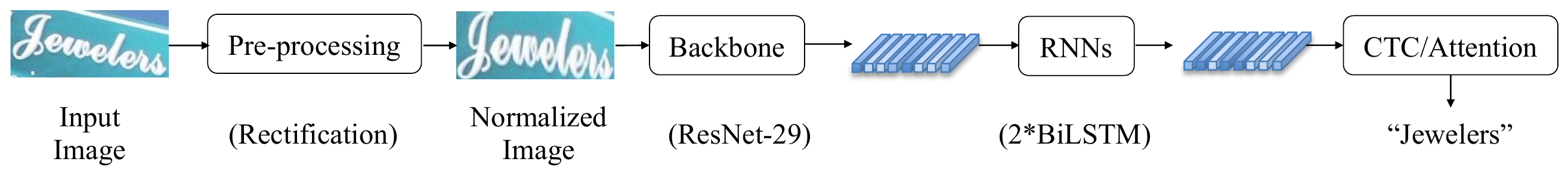}
  \caption{Architecture of the recognizer~\cite{aberdam2021sequence,baek2019wrong}.  
}
  \label{pic-reg}
  \vspace{-1.em}
\end{figure*}

\subsection{Dataset}
We evaluate our approach on several public benchmarks that are widely used in scene text recognition studies. These datasets include \textbf{IC03}~\cite{lucas2003icdar}, \textbf{IC13}~\cite{karatzas2013icdar}, \textbf{IC15}~\cite{karatzas2015icdar}, \textbf{SVT}~\cite{wang2011end}, \textbf{SVT-P}~\cite{quy2013recognizing}, \textbf{IIIT5K}~\cite{mishra2012scene}, CUTE80 (\textbf{CT80})~\cite{risnumawan2014robust} and Total-Text (\textbf{TText})~\cite{ch2020total}.

We construct a dataset for self-supervised representation learning. To obtain more realistic and diverse scene text images, we collect samples from public real training datasets, including IIIT5K~\cite{mishra2012scene}, IC13~\cite{karatzas2013icdar}, IC15~\cite{karatzas2015icdar}, COCO-Text~\cite{veit2016coco}, RCTW~\cite{shi2017icdar2017}, ArT~\cite{chng2019icdar2019}, ReCTS~\cite{zhang2019icdar}, MTWI~\cite{he2018icpr2018}, LSVT~\cite{sun2019chinese} and MLT~\cite{nayef2019icdar2019}. We discard low-resolution images with a height of less than 32 pixels or width of less than 64 pixels (the width should be greater than two times the height for constructing training samples). Because in practice, low-quality images confuse the image recovery task and lead to inefficient training. As a result, we discard their labels and obtain an unlabeled dataset composed of approximately 300k real samples, termed \textbf{Real-300K}\footnote{\url{https://github.com/Canjie-Luo/Real-300K}.}. Besides, we also use the popular synthetic dataset \textbf{SynthText}~\cite{gupta2016synthetic} for fair comparisons with the previous study SeqCLR~\cite{aberdam2021sequence}.

\subsection{Implementation Details}
We provide more details, such as augmentations, architectures, probe objectives, and training settings, in the \textsl{Supplementary Material}.

\textbf{Encoder/Decoder}  We adopt a popular recognizer backbone ResNet-29~\cite{baek2019wrong} as our encoder. We symmetrically design a lightweight decoder. 

\textbf{Recognizer} The complete architecture of the recognizer follows~\cite{aberdam2021sequence,baek2019wrong}, including a rectification module, a ResNet-29 backbone, two stacked BiLSTMs and a CTC~\cite{graves2006connectionist} /Attention~\cite{bahdanau2014neural} decoder, as shown in Figure~\ref{pic-reg}.

\textbf{Optimization} In the self-supervised representation learning stage, we set the batch size to 256 and train the network for 400K iterations. It takes less than 3 days for convergence on two NVIDIA P100 GPUs (16GB memory per GPU). The optimizer is Adam~\cite{kingma2014adam} with the settings of $\beta_{1}=0.5$ and $\beta_{2}=0.999$. The learning rate is set to $10 ^{-4}$ and linearly decreased to $10^{-5}$. The images are resized to a height of 32 pixels, maintaining the aspect ratio. The training setting of recognizers follows previous study SeqCLR~\cite{aberdam2021sequence}.

\begin{table*}[t]
\centering
\begin{minipage}[!t]{1.4\columnwidth}
\caption{Probe evaluation. We report the word-level accuracy (Acc., \%) and accuracy up to one edit distance (E.D. 1, \%). Although we cannot perform direct comparisons with SeqCLR, we list its results for reference. The ``Proj.", ``Seq. Map.", ``Att." denotes projection head, sequential mapping, and attention, respectively. The RNN is a BiLSTM (256 hidden units).}
\centering
\begin{adjustbox}{width=1\textwidth}
\begin{tabular}{ c c c c c c c c c c}
\toprule
\multirow{2}{*}{Method} & \multirow{2}{*}{Encoder} & Decode Block & Probe & \multicolumn{2}{c}{IIIT5K} & \multicolumn{2}{c}{IC03} & \multicolumn{2}{c}{IC13} \\ \cmidrule(lr){5-6} \cmidrule(lr){7-8} \cmidrule(lr){9-10}
 & & (Train) & (Test) & Acc. & E.D. 1 & Acc. & E.D. 1 & Acc. & E.D. 1 \\ 
\midrule
\midrule
SeqCLR~\cite{aberdam2021sequence} & ResNet + 2*RNN & Proj. + Seq. Map. & CTC & {35.7} & {62.0} & {43.6} & {71.2} & {43.5} & {67.9} \\
\midrule
\multirow{4}{*}{Ours} & ResNet + 2*RNN & FCN & CTC & 0.0 & 2.8 & 0.0 & 0.0 & 0.0 & 6.4 \\
 & ResNet & FCN & CTC & \textbf{1.5} & \textbf{7.9} & \textbf{2.3} & \textbf{5.2} & \textbf{2.2} & \textbf{12.9} \\
 \cmidrule(lr){2-10}
 & ResNet & FCN & 1*RNN + CTC & 57.4 & 75.1 & 64.8 & \textbf{78.9} & 63.0 & \textbf{81.2} \\
 & ResNet & FCN & 2*RNN + CTC & \textbf{60.8} & \textbf{75.6} & \textbf{64.9} & \textbf{78.9} & \textbf{64.0} & 81.0 \\
\midrule
\midrule
SeqCLR~\cite{aberdam2021sequence} & ResNet + 2*RNN & Proj. + Seq. Map. & Att. & {49.2} & {68.6} & {63.9} & {79.6} & {59.3} & {77.1} \\
\midrule
\multirow{4}{*}{Ours} & ResNet + 2*RNN & FCN & Att. & 6.4 & 12.8 & 6.8 & 9.9 & 7.1 & 15.1 \\
 & ResNet & FCN & Att. & \textbf{22.2} & \textbf{39.7} & \textbf{22.3} & \textbf{38.6} & \textbf{24.1} & \textbf{43.6} \\
 \cmidrule(lr){2-10} 
 & ResNet & FCN & 1*RNN + Att. & 65.0 & 78.3 & \textbf{73.6} & \textbf{85.9} & \textbf{71.8} & \textbf{84.3} \\
 & ResNet & FCN & 2*RNN + Att. & \textbf{66.5} & \textbf{78.8} & 71.7 & 83.6 & 68.7 & 81.6 \\
\bottomrule
\end{tabular}
\end{adjustbox}
\label{tab-probe}
\end{minipage}
\ \ 
\begin{minipage}[!t]{0.65\columnwidth}
\caption{Comparisons of augmentation strategies. We discard the spatial transformation augmentations because our approach recovers images based on consistent visual cues.}
\centering
\begin{adjustbox}{width=1\textwidth}
\begin{tabular}{c c c}
\toprule
\multirow{2}{*}{Aug. Strategy} & Contrastive & Generative \\ 
 & (SeqCLR~\cite{aberdam2021sequence}) & (Ours) \\ 
\midrule
Color Contrast & \checkmark & \checkmark \\
\midrule
Blurring & \checkmark & \checkmark \\
\midrule
Sharpen Blending & \checkmark & \checkmark \\
\midrule
Random Noise & \checkmark & \checkmark \\
\midrule
Cropping & \checkmark & $\times$ \\
\midrule
Perspective Trans. & \checkmark & $\times$ \\
\midrule
Piecewise Affine & \checkmark & $\times$ \\
\bottomrule
\end{tabular}
\end{adjustbox}
\label{tab-aug}
\end{minipage}
\vspace{-1.em}
\end{table*}

\subsection{Probe Evaluation}
We first study the representation quality using the common protocol, namely probe evaluation. Specifically, we perform self-supervised pre-training of the ResNet-29 backbone using SynthText~\cite{gupta2016synthetic}. Then we fix the parameters of the backbone and feed the frozen representations to a CTC/Attention probe. The probes are trained on the same labeled SynthText dataset. It is believed that the higher the representation quality, the better the probe can obtain cues for classification. 

The quantized results, including word accuracy (Acc.) and word-level accuracy up to one edit distance (E.D. 1)~\cite{aberdam2021sequence}, are reported in Table~\ref{tab-probe}. Note that our generative scheme is significantly different from the contrastive scheme SeqCLR~\cite{aberdam2021sequence}, which uses sufficient sequential modeling (RNN projection head and sequential mapping) in the self-supervised pre-training phase. Although the direct comparisons between the two approaches are somewhat unreasonable, we list SeqCLR's results under a similar experimental setting for reference.

Here we analyze the results of our approach. Note that the sequential modeling (2*RNN) in the encoder reduces the quality of representations. This is because our approach models local patterns for recovery, but the sequential modeling introduces contexts to disturb this learning scheme. Therefore, we discard the sequential modeling in the encoder. This means our approach might lack the capacity of sequence modeling after self-supervised representation learning. However, it is possible to equip a lightweight RNN in the probe, which remarkably improves the representation quality. Overall, we obtain promising representations in a generative manner. This might bring a brand new learning perspective in the field of scene text recognition.

Moreover, we find that this experimental setting (pre-training the backbone and fine-tuning the probe using the very same synthetic dataset) might not meet the actual practice. In fact, we usually encounter one situation that we have vast amounts of unlabeled real-world data. It is worth making the best use of the real-world data. Therefore, we conduct an experiment under this new setting to further verify the effectiveness of our approach. We perform self-supervised learning of the backbone using the Real-300K dataset. As shown in Table~\ref{tab-probe-real}, the recognition performance is significantly boosted. As the real-world dataset provides more realistic and diverse images, it benefits the robustness of the backbone. Another reason why using a real dataset achieves better results might be the closer distribution to the benchmarks, which are also real-world datasets.

\begin{table}[t]
\centering
\caption{Probe evaluation. We report the word accuracy (Acc., \%) and word-level accuracy up to one edit distance (E.D. 1, \%). The real training data provides more robust representations.}
\begin{adjustbox}{width=0.48\textwidth}
\begin{tabular}{c c c c c c c c c}
\toprule
Probe & \multicolumn{2}{c}{Training Data} & \multicolumn{2}{c}{IIIT5K} & \multicolumn{2}{c}{IC03} & \multicolumn{2}{c}{IC13} \\ 
 \cmidrule(lr){2-3}  \cmidrule(lr){4-5}  \cmidrule(lr){6-7}  \cmidrule(lr){8-9} 
Type& Encoder & Probe & Acc. & E.D. 1 & Acc. & E.D. 1 & Acc. & E.D. 1 \\ 
\midrule
\multirow{2}{*}{CTC} & Synth. & Synth. & 60.8 & 75.6 & 64.9 & 78.9 & 64.0 & 81.0 \\
& Real & Synth. & \textbf{68.9} & \textbf{82.8} & \textbf{75.0} & \textbf{87.2} & \textbf{72.9} & \textbf{86.0} \\ 
\midrule
\multirow{2}{*}{Att.} & Synth. & Synth. & 66.5 & 78.8 & 71.7 & 83.6 & 68.7 & 81.6 \\
& Real & Synth. & \textbf{73.7} & \textbf{85.6} & \textbf{81.2} & \textbf{90.4} & \textbf{77.9} & \textbf{87.8} \\ 
\bottomrule
\end{tabular}
\end{adjustbox}
\label{tab-probe-real}
\vspace{-1.2em}
\end{table}

\textbf{Discussion} Here we reveal two significant differences between the contrastive learning scheme SeqCLR and our generative learning scheme SimAN. 1) We summarize the augmentation strategies in Table~\ref{tab-aug}. As our SimAN recovers an image according to the consistent visual cues, we do not introduce spatial transformation augmentations into our pipeline. This means that our approach is more suitable for scene text images, rather than handwritten text images (focusing on stroke deformations) in black and white. On the contrary, the SeqCLR shows more promising results on handwritten text than scene text. 2) We find that adding a sequence model in the encoder yields degraded performance of our approach, whereas it provides noteworthy improvements for SeqCLR. This is because our approach models local patterns for recovery, while the SeqCLR requires contextual information within the sequence for discrimination. 

There exist different properties of the two schemes. In this regard, the complementarity of contrastive and generative approaches is worth future explorations.

\subsection{Semi-Supervision Evaluation}
We further study the performance under a semi-supervision manner. Since it can make the best use of abundant unlabeled data, it has important practical significance. As SynthText provides six million training samples, it is able to sample smaller subsets with three orders of scales (10K, 100K, and 1M from the original 6M data). After performing self-supervised pre-training of the backbone on SynthText, we use the pre-trained parameters to initialize the recognizer backbone. Finally, we fine-tune the entire recognizer using different subsets of SynthText. 

\begin{table*}[h]
\centering
\caption{Semi-supervised performance evaluation. We sample three orders of scales (10K, 100K, and 1M) of data from SynthText (6M). Our approach can learn high-quality representations from unlabeled data and improve the supervised baseline, especially when used with low-resource labeled data.}
\begin{adjustbox}{width=1\textwidth}
\begin{tabular}{ c c llll llll llll}
\toprule
\multirow{3}{*}{Method} & \multirow{3}{*}{Supervision} & \multicolumn{4}{c}{IIIT5K} & \multicolumn{4}{c}{IC03} & \multicolumn{4}{c}{IC13} \\ 
\cmidrule(lr){3-6} \cmidrule(lr){7-10} \cmidrule(lr){11-14}
&  & \multicolumn{4}{c}{Labeled Training Data} & \multicolumn{4}{c}{Labeled Training Data} & \multicolumn{4}{c}{Labeled Training Data}  \\
&  & 10K & 100K & 1M & 6M & 10K & 100K & 1M & 6M & 10K & 100K & 1M & 6M \\
\midrule
\multirow{2}{*}{SeqCLR~\cite{aberdam2021sequence}} & Sup. & - & - & - & \textbf{83.8} & - & - & - & 91.1 & - & - & - & \textbf{88.1} \\
                  & Semi-Sup. & - & - & - & 82.9 \textbf{\textcolor{red}{$\downarrow$ 0.9}} & - & - & - & \textbf{92.2} \textbf{\textcolor{myGreen}{$\uparrow$ 1.1}} & - & - & - & 87.9 \textbf{\textcolor{red}{$\downarrow$ 0.2}} \\ 
\midrule
\multirow{2}{*}{Ours} & Sup. & 35.0 & 72.6 & \textbf{84.1} & 86.6 & 37.6 & 79.4 & 88.2 & 91.5 & 38.6 & 75.3 & 86.4 & 89.0 \\
                  & Semi-Sup. & \textbf{41.1} \textbf{\textcolor{myGreen}{$\uparrow$ 6.1}} & \textbf{73.6} \textbf{\textcolor{myGreen}{$\uparrow$ 1.0}} & \textbf{84.1} & \textbf{87.5} \textbf{\textcolor{myGreen}{$\uparrow$ 0.9}} & \textbf{42.9} \textbf{\textcolor{myGreen}{$\uparrow$ 5.3}} & \textbf{79.9} \textbf{\textcolor{myGreen}{$\uparrow$ 0.5}} & \textbf{89.2} \textbf{\textcolor{myGreen}{$\uparrow$ 1.0}} & \textbf{91.8} \textbf{\textcolor{myGreen}{$\uparrow$ 0.3}} & \textbf{43.9} \textbf{\textcolor{myGreen}{$\uparrow$ 5.3}} & \textbf{75.6} \textbf{\textcolor{myGreen}{$\uparrow$ 0.3}} & \textbf{86.5} \textbf{\textcolor{myGreen}{$\uparrow$ 0.1}} & \textbf{89.9} \textbf{\textcolor{myGreen}{$\uparrow$ 0.9}} \\ 
\bottomrule
\end{tabular}
\end{adjustbox}
\label{tab-semi}
\vspace{-1.em}
\end{table*}

As shown in Table~\ref{tab-semi}, our approach using the semi-supervised setting outperforms the supervised baseline. For instance, under the 10K low-resource setting, our approach increases the accuracy by more than 5\%, which suggests that the recognition robustness is highly correlated with representation quality. With the increase of the scale of labeled data, our approach can still contribute to recognition accuracy. We compare the semi-supervised results with the previous study termed SeqCLR~\cite{aberdam2021sequence} under the same setting. Note that our approach can still slightly improve recognition performance using the whole SynthText for fine-tuning, whereas the SeqCLR shows inconsistent performance. This indicates the generalization ability of our approach.

\subsection{Generative Visual Tasks}
We demonstrate the potential of our approach on generative visual effect tasks. For the generalization to several different tasks, we adopt a widely used VGG encoder and a corresponding decoder~\cite{huang2017arbitrary,johnson2016perceptual} in our model. The training dataset is Real-300K. The image height is set to 64 pixels. 

\subsubsection{Data Synthesis}
\begin{figure}[h]
  \centering
  \includegraphics[width=1\columnwidth]{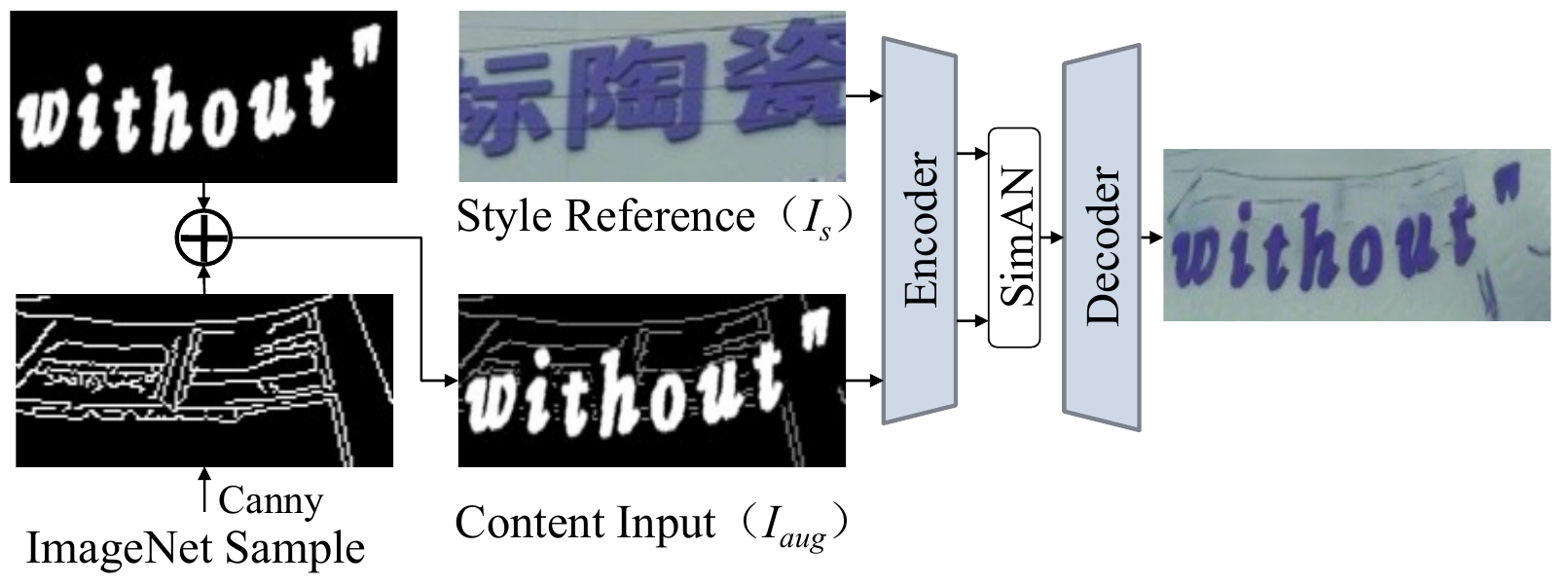}
  \caption{Pipeline of data synthesis. We can synthesize similar style images containing new text strings. Note that the sketch on the canvas $I_{aug}$ is also aligned with corresponding style of background noise on the source image $I_s$.}
  \label{pic-synth-pipeline}
  \vspace{-1.2em}
\end{figure}

As our generative learning scheme decouples content and style representations, we can randomly integrate existing styles and new contents to synthesize diverse training samples. As shown in Figure~\ref{pic-synth-pipeline}, we replace the $I_s$ with a style reference image and replace the $I_{aug}$ with a new content input. Then the generative network can synthesize an image in a similar style retaining the required content. Note that the terms ``style" and ``content" are somewhat different from those of font style transfer tasks~\cite{wang2021deepvecfont}. Here the style refers to aspects such as the color, blurring level, and textures, rather than the font category. The term content indicates not only the text string but also the outline of backgrounds and the topological shape of fonts. Thus, it is possible to introduce more background noise by adding variant sketches extracted by the Canny edge detection operator on ImageNet samples~\cite{krizhevsky2012imagenet}. Thus, a clean canvas containing a slanted/curved text can be finally rendered as abundant diverse scene text images. 

\begin{figure}[h]
  \centering
  \includegraphics[width=1\columnwidth]{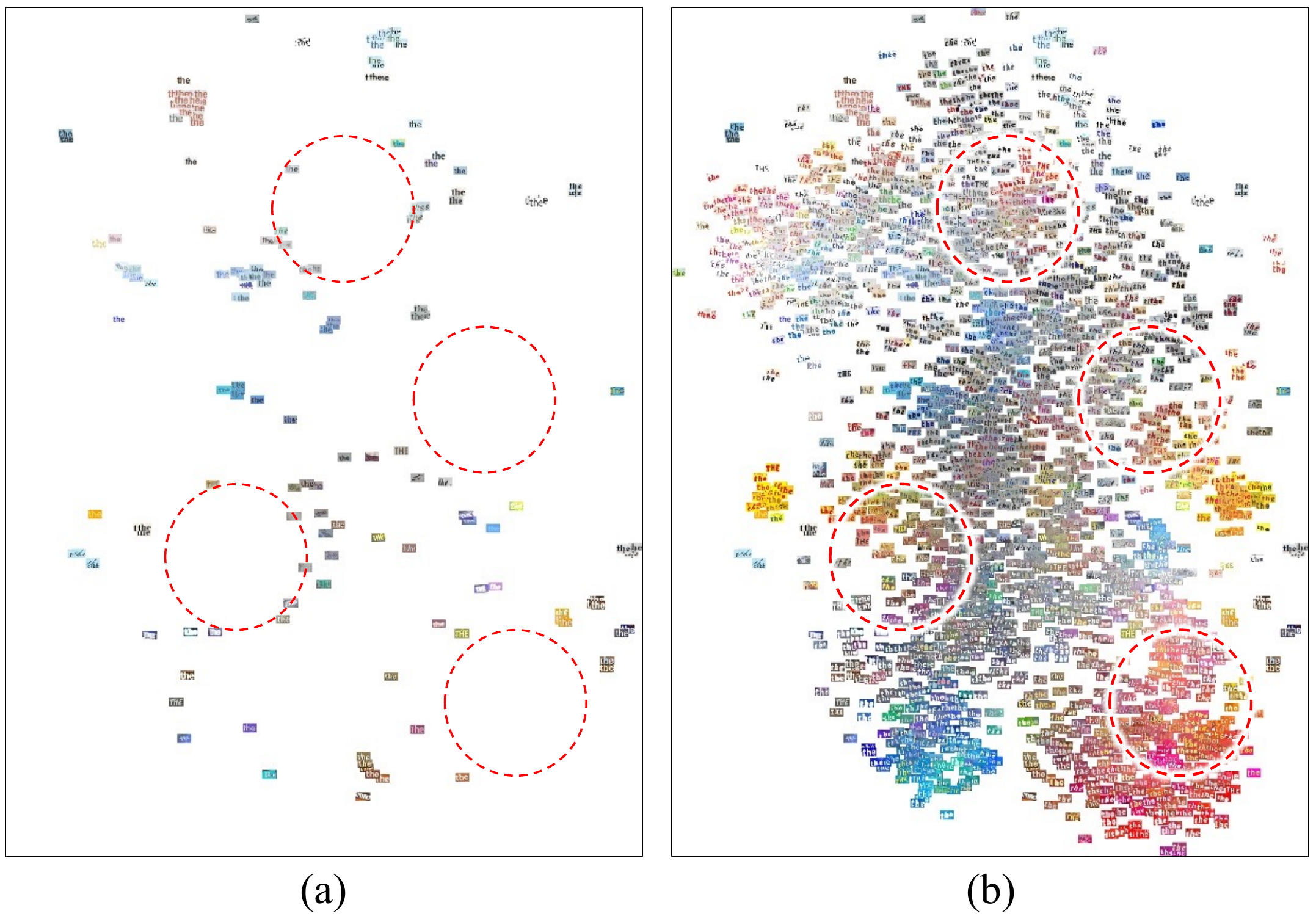}
  \caption{Distribution of scene text images containing the word ``the" via t-SNE. We show two distributions of (a) 200 real labeled samples and (b) 200 real samples and our 2000 synthetic samples. The large empty space of original distribution might suggest the lack of diversity of labeled data. After adding our synthetic samples, the distribution is more even and dense. Best viewed in color.}
  \label{pic-synth-dist}
  \vspace{-1.2em}
\end{figure}

\begin{table}[h]
\caption{Word accuracy (\%) on benchmarks. Following the UnrealText~\cite{long2020unrealtext}, we synthesize 1M samples and train the same recognizer. For each column, the best result is highlighted in \textbf{bold} font, and the second-best result is shown with an \underline{underline}.}
\centering
\begin{adjustbox}{width=0.46\textwidth}
\begin{tabular}{c c c c c c c}
\toprule
 Method & IIIT5K & SVT & IC15 & SVT-P & CT80 & TText \\
 \midrule
Synth90K~\cite{jaderberg2016reading} & 51.6 & 39.2 & 35.7 & 37.2 & 30.9 & 30.5 \\
SynthText~\cite{gupta2016synthetic} & 53.5 & 30.3 & 38.4 & 29.5 & 31.2 & 31.1 \\
 Verisimilar Synthesis~\cite{zhan2018verisimilar} & 53.9 & 37.1 & 37.1 & 36.3 & 30.5 & 30.9 \\
 UnrealText~\cite{long2020unrealtext} & 54.8 & 40.3 & \textbf{39.1} & \underline{39.6} & 31.6 & 32.1 \\
 \midrule
 Ours (high res., 64$\times$) & \underline{62.3} & \underline{51.2} & 35.0 & 36.6 & \underline{44.8} & \underline{37.9} \\
 Ours (blurred) & \textbf{65.7} & \textbf{58.6} & \underline{38.7} & \textbf{44.2} & \textbf{47.9} & \textbf{38.3} \\
\bottomrule
\end{tabular}
\end{adjustbox}
\label{tab-reg}
\vspace{-1.5em}
\end{table}

First, we visualize the distributions of the limited real labeled samples and our plentiful synthetic samples. As shown in Figure~\ref{pic-synth-dist}, the limited labeled real-world data cannot cover diverse styles. However, our synthetic data fills the empty style space, indicating the significantly enriched styles. Then, we conduct recognition experiments to show the quantitative results. Following the settings of UnrealText~\cite{long2020unrealtext}, we synthesize 1M samples to train the same recognizer and report the accuracy on several benchmarks. As shown in the second last row in Table~\ref{tab-reg}, our samples outperform previous synthesis methods~\cite{jaderberg2016reading,gupta2016synthetic,zhan2018verisimilar,long2020unrealtext} on four (out of six) benchmarks without bells and whistles. We find that our synthetic samples have a high resolution (height of 64 pixels), which usually cannot meet the low-quality practice of scene text. Therefore, we simply add blurring to the samples. The recognition performance is further boosted, suggesting that our synthesis pipeline is scalable. 

\vspace{-.6em}
\subsubsection{Arbitrary-Length Text Editing}
\begin{figure}[h]
  \centering
  \includegraphics[width=0.9\columnwidth]{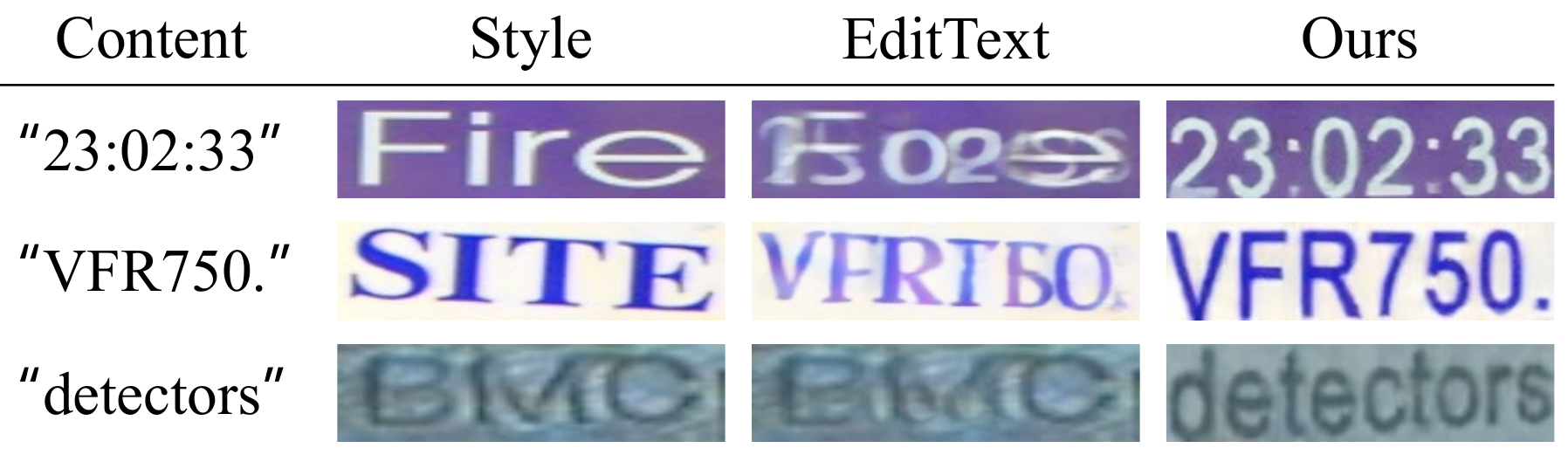}
  \vspace{-0.8em}
  \caption{Visualization of text editing. The EditText~\cite{wu2019editing} cannot deal with target strings of variant lengths, whereas our approach adaptively aligns correct styles and achieves more readable results.}
  \label{pic-edit}
  \vspace{-1.5em}
\end{figure}

\begin{figure}[t]
  \centering
  \includegraphics[width=0.85\columnwidth]{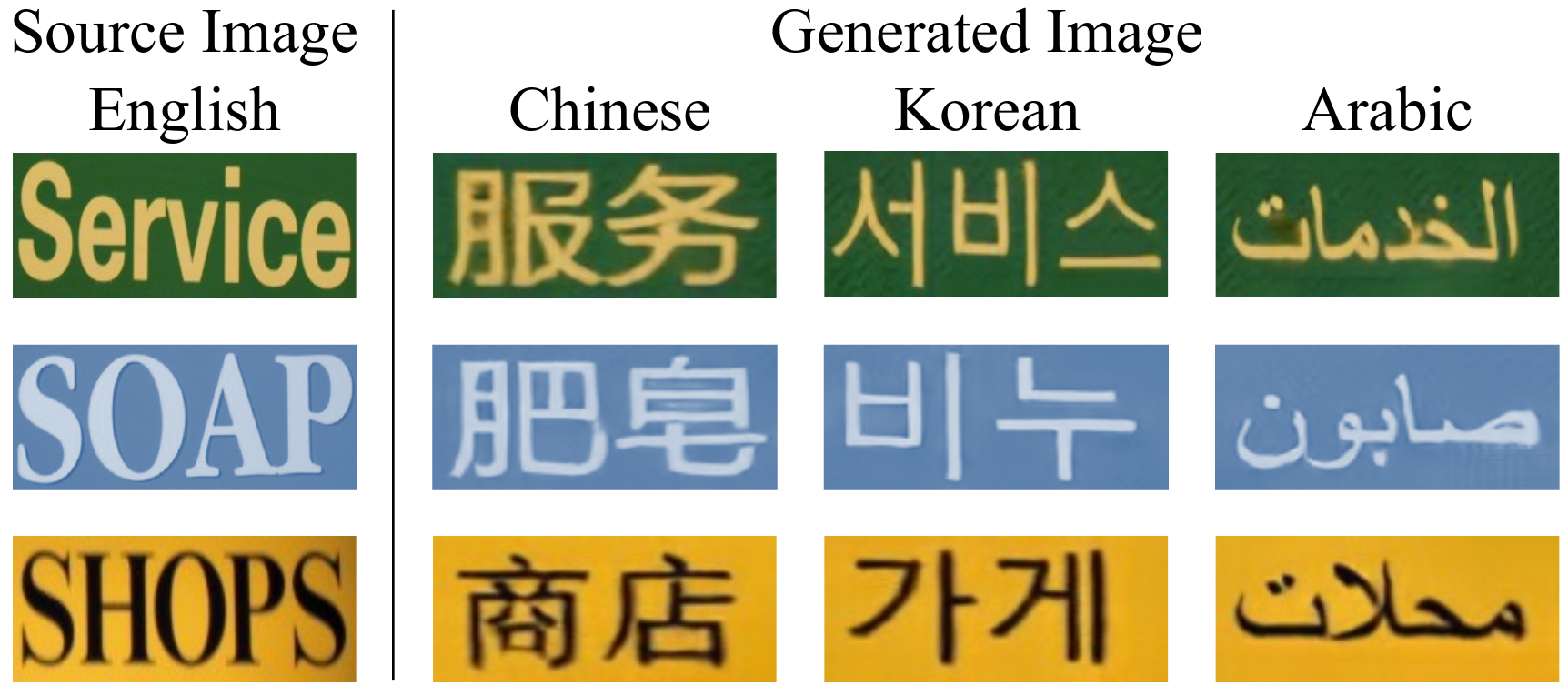}
  \caption{Cross language editing via our self-supervised approach.}
  \label{pic-cross-lang}
  \vspace{-0.5em}
\end{figure}

\vspace{-.3em}
The goal of editing text in the wild is to change the word on the source image while retaining the realistic source look. As our approach can synthesize new words within source styles, we study the performance of our self-supervised approach and a popular supervised method EditText\footnote{\url{https://github.com/youdao-ai/SRNet}}~\cite{wu2019editing}. We generate 10K images using the corpus of SynthText~\cite{gupta2016synthetic} and the style of IC13~\cite{karatzas2013icdar}. Then we evaluate the style distribution similarity using the FID score~\cite{heusel2017gans} and the readability using a mainstream recognizer\footnote{\url{https://github.com/meijieru/crnn.pytorch}}~\cite{shi2017end}. As shown in Figure~\ref{pic-edit} and Table~\ref{tab-edit}, the EditText cannot handle target text of various lengths. That means the editing is limited to approximately the same length words. Although its style distribution is closer to the source images, its generated images are unreadable. On the contrary, our approach can adaptively align correct styles to arbitrary-length text, indicating the flexibility of our self-supervised approach. In our practice, we find that our approach is sufficient for cross-language editing, as shown in Figure~\ref{pic-cross-lang}. It has a wide range of applications, such as menu translation and cross-border e-commerce.
\vspace{-0.5em}

\begin{table}[h]
\centering
\caption{Arbitrary-length Text editing evaluation. We report FID score and word-level recognition accuracy (\%). Although the supervised EditText can imitate more font category and background texture, our self-supervised approach achieves better readability.}
\begin{adjustbox}{width=0.32\textwidth}
\begin{tabular}{c c c c}
\toprule
Method & Supervision & FID $\downarrow$ & Acc. $\uparrow$ \\
\midrule
EditText~\cite{wu2019editing} & \checkmark & \textbf{40.5} & 14.9 \\
Ours & $\times$ & 67.9 & \textbf{57.6} \\
\bottomrule
\end{tabular}
\end{adjustbox}
\label{tab-edit}
\vspace{-1.6em}
\end{table}

\subsubsection{Font Interpolation}
\begin{figure}[t]
  \centering
  \includegraphics[width=0.95\columnwidth]{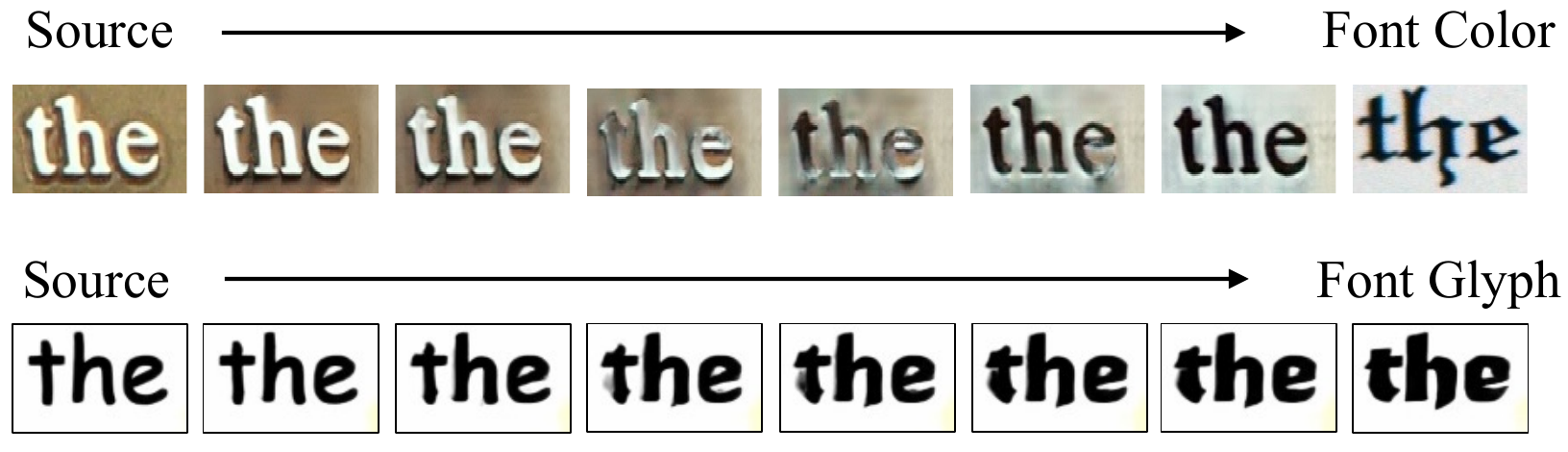}
  \caption{Font interpolation effects produced by our approach.}
  \label{pic-font}
  \vspace{-1.5em}
\end{figure}

\vspace{-.4em}
It is believed that font design is a professional technique belonging to a few experts~\cite{wang2021deepvecfont}. We present an interesting application of our approach on font interpolation for automatically and efficiently generating font candidates. As we parameterize the style and content as representations, we can interpolate these representations to achieve transitional effects. For instance, we compute the style representations (local statistics) of two images and rearrange them according to the same content representations. We interpolate the two style representations to decode images so that we can obtain the gradually changing colors, sheens, and shadows, as shown in Figure~\ref{pic-font}. Simultaneously, we interpolate the content representations to achieve font glyph changes. This potential suggests our approach might facilitate font design.

\vspace{-1.2em}
\section{Broader Impacts}
\vspace{-.4em}
The proposed self-supervised approach has a wide range of applications owing to its capability of decoupling styles and contents of scene text. For instance, it can swap text to achieve image (and video) manipulation, which can be used in many applications, such as menu translation and cross-border e-commerce. However, we point out the risks of text image editing. It can be employed to tamper sensitive data, such as personal information, license plate numbers, and financial statistics, to trick systems that rely on text recognition. It is necessary to reduce these negative impacts. One promising technological solution is to detect the edited/attacking image using a qualified discriminator. It is also essential to increase media literacy among vast swathes of the population.

\section{Conclusion}
\vspace{-.35em}
We have presented a novel approach for self-supervised representation learning of scene text from a brand new perspective, \textit{i.e.}, in a generative manner. It takes advantage of the style consistency of neighboring patches among one text image to reconstruct one augmented patch under the guidance of its neighboring patch. Specifically, we propose a SimAN module to identify different patterns (\textit{e.g.}, background noise and foreground characters) based on the representation similarity between the two patches. The representations are required to be sufficiently distinguishable so that corresponding styles can be correctly aligned to reconstruct the augmented patch. Otherwise, it results in an inaccurate image. In this way, it enables self-supervised representation learning via the image reconstruction task.

Extensive experiments show that our generative approach achieves promising representation quality and outperforms the previous contrastive method. Furthermore, it presents the impressive potential for data synthesis, text image editing and font interpolation, demonstrating a wide range of practical applications. Our study might arouse the rethinking of self-supervised learning of scene text. In the future, we will study the complementarity of contrastive and generative learning schemes to further improve the representation quality.

\vspace{-.35em}
\section*{Acknowledgment}
\vspace{-.35em}
This research was supported in part by NSFC (Grant No. 61936003) and GD-NSF (No. 2017A030312006).

\clearpage
{\small
\bibliographystyle{ieee_fullname}
\bibliography{egbib}
}

\setcounter{table}{0}
\setcounter{figure}{0}
\setcounter{section}{0}

\twocolumn[{
\renewcommand\twocolumn[1][]{#1}
\centering
\Large
\textbf{SimAN: Exploring Self-Supervised Representation Learning of Scene Text \\
via Similarity-Aware Normalization} \\
\vspace{0.5em}Supplementary Material \\
\vspace{1.0em}
\renewcommand{\arraystretch}{2}
\captionof{table}{Visualization of the self-supervised learning scheme. The queries are denoted as red boxes. The proposed SimAN requires distinguishable representations to identify different patterns, thus enabling self-supervised representation learning of the encoder. Under the supervision of the $\mathcal{L}_{2}$, the responses on the neighboring patches are becoming more and more accurate, suggesting the increasing quality of the representations.} 
\vspace{-1.5em}
\begin{center}
\begin{adjustbox}{width=1\textwidth}
\begin{tabular}{c c l}
\toprule
\multicolumn{2}{c}{Query} & \begin{minipage}{1.5\textwidth} \includegraphics[width=1.\columnwidth]{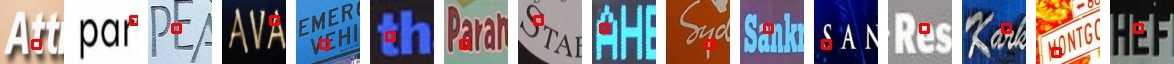} \end{minipage} \\ 
\multicolumn{2}{c}{Key (neighboring patch)} & \begin{minipage}{1.5\textwidth} \includegraphics[width=1.\columnwidth]{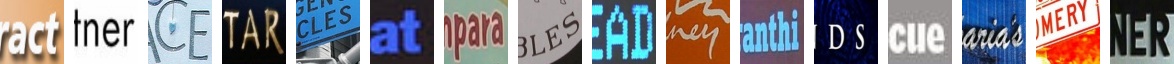} \end{minipage} \\
\midrule
\multirow{3.6}{*}{Mask} & $\mathcal{L}_{2}\approx 0.3$ & \begin{minipage}{1.5\textwidth} \includegraphics[width=1.\columnwidth]{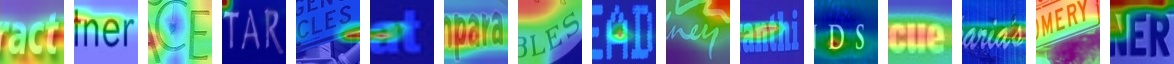} \end{minipage}\\
& $\mathcal{L}_{2}\approx 0.1$ & \begin{minipage}{1.5\textwidth} \includegraphics[width=1.\columnwidth]{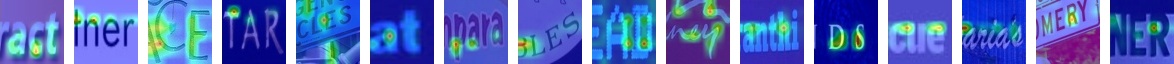} \end{minipage}\\
& $\mathcal{L}_{2}\approx 0.02$ & \begin{minipage}{1.5\textwidth} \includegraphics[width=1.\columnwidth]{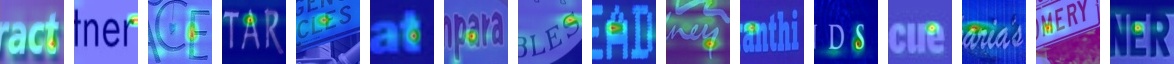} \end{minipage}\\
\bottomrule 
\end{tabular}
\end{adjustbox}
\end{center}
\label{tab-visual}
\vspace{-0em}
}]

\setcounter{page}{1}

\section{Visualization}
To validate the effectiveness of the proposed SimAN, which estimates pattern similarity and then queries corresponding style keys for recovering the augmented patch, we visualize the attentional responses of style keys on the neighboring patch. As shown in Table~\ref{tab-visual}, with the decrease of the $\mathcal{L}_{2}$ loss, the attentional mask presents a more and more accurate response based on a similar pattern. For instance, as shown in the first column, a ``t'' query (denoted as a red box) on the source patch obtains a response of ``t'' on the neighboring patch. This reveals the learning mechanism of the proposed SimAN, \textit{i.e.}, distinguishable representations between different characters are required to identify patterns and align correct styles for image reconstruction.

\section{Benchmark}
We detail the public scene text benchmarks used for recognition evaluation as follows.

ICDAR 2003~\cite{lucas2003icdar} (\textbf{IC03}) contains 867 cropped images after discarding images that contain non-alphanumeric characters or less than three characters~\cite{wang2011end}. 

ICDAR 2013~\cite{karatzas2013icdar} (\textbf{IC13}) inherits most of its samples from IC03. It contains 1015 cropped images.

ICDAR 2015~\cite{karatzas2015icdar} (\textbf{IC15}) was collected by using Google Glasses. It includes more than 200 irregular text images.

Street View Text~\cite{wang2011end} (\textbf{SVT}) consists of 647 word images for testing. Some images are severely corrupted by noise and blur. 

Street View Text Perspective~\cite{quy2013recognizing} (\textbf{SVT-P}) is a perspective distorted version of \textbf{SVT}, containing 645 cropped images for testing. 

IIIT5K-Words~\cite{mishra2012scene} (\textbf{IIIT5K}) contains 3000 and 2000 cropped word images for testing and training, respectively. Some texts are curved.

CUTE80~\cite{risnumawan2014robust} (\textbf{CT80}) was specifically collected to evaluate the performance of curved text recognition. It contains 288 cropped natural images.

Total-Text~\cite{ch2020total} (\textbf{TText}) focuses on curved text recognition. It contains 2201 cropped word images.

\section{Augmentation Strategy}
Different from the previous study SeqCLR~\cite{aberdam2021sequence}, we discard the spatial transformation augmentations because our approach recovers images based on consistent visual cues. Therefore, we limit the augmentation strategies to color changes, blurring, sharpen blending, and random noise. We use a CPU-efficient toolkit\footnote{ \small \url{https://github.com/albumentations-team/albumentations}} to perform augmentation. The pseudo-code is shown as below for reference. 

\begin{lstlisting}[language=Python]
import albumentations as A
A.Sequential([
    # Color Changes
    A.InvertImg(),
    A.OneOf([
        A.ChannelDropout(),
        A.ChannelShuffle(),
        A.ToGray(),
        A.RGBShift(),
        A.Equalize(),
        A.RandomBrightnessContrast(0.5, 0.5),
        A.ColorJitter(0.5, 0.5, 0.5, 0.5),
        A.HueSaturationValue(),
        A.RandomToneCurve(),
    ]),
    A.OneOf([
        # Sharpen Blending
        A.Sharpen(alpha=(1.0, 1.0)),
        # Blurring
        A.OneOf([
            A.ImageCompression(40, 80),
            A.Blur(blur_limit=[3, 3]),
            A.GaussianBlur(blur_limit=[3, 3]),
            A.MedianBlur(blur_limit=[3, 3]),
            A.MotionBlur(blur_limit=[3, 3]),
        ]),
        # Random Noise
        A.OneOf([
            A.Emboss((0.5, 1.0), (0.8, 1.0)),
            A.GaussNoise(),
            A.ISONoise((0.1, 0.5), (0.5, 1.0)),
            A.MultiplicativeNoise(),
        ]),
    ]),
])
\end{lstlisting}

\begin{table}[h]
\centering
\caption{Probe evaluation using an attentional probe with two BiLSTMs (256 hidden units). We report the word accuracy (Acc., \%) and word-level accuracy up to one edit distance (E.D. 1, \%). The two augmentation toolkits achieve comparable performance.}
\begin{adjustbox}{width=0.45\textwidth}
\begin{tabular}{c c c c c c c}
\toprule
Augmentation & \multicolumn{2}{c}{IIIT5K} & \multicolumn{2}{c}{IC03} & \multicolumn{2}{c}{IC13} \\ 
 \cmidrule(lr){2-3}  \cmidrule(lr){4-5}  \cmidrule(lr){6-7}
Toolkit & Acc. & E.D. 1 & Acc. & E.D. 1 & Acc. & E.D. 1 \\ 
\midrule
SeqCLR's & 65.6 & 78.2 & 71.3 & \textbf{84.2} & \textbf{69.4} & \textbf{82.2} \\
Ours & \textbf{66.5} & \textbf{78.8} & \textbf{71.7} & 83.6 & 68.7 & 81.6 \\ 
\bottomrule
\end{tabular}
\end{adjustbox}
\label{tab-toolkit}
\vspace{-0.5em}
\end{table}

Note that we use a different toolkit \textsl{albumentations} from that of SeqCLR~\cite{aberdam2021sequence}. We clarify the performance gain is achieved by our proposed approach, rather than the different toolkit. As shown in Table~\ref{tab-toolkit}, the two augmentation toolkits achieve comparable performance.

\begin{table}[t]
\caption{Architecture of ResNet-29. We present the size of feature maps during representation learning and recognition training. The backbone (encoder) trained on image patches can generalize well to images of variant widths. We pad the output feature maps (whose height is one) along the vertical direction for the extraction of eight-neighborhood statistics. } 
\begin{centering}
\begin{adjustbox}{width=0.43\textwidth}
\begin{tabular}[t]{c lc c c}
\toprule
\multirow{2}{*}{\textbf{Layers}} & \multicolumn{2}{c}{\multirow{2}{*}{\textbf{Configurations}}} & \multicolumn{2}{c}{\textbf{Size}} \\
\cmidrule(lr){4-5}
& \multicolumn{2}{c}{} & Repr. Learn. & Reg. \\
\midrule
Input & \multicolumn{2}{c}{RGB image} & $32\times 32$ & $32\times 100$\\
\midrule
Conv1 & c: $32$ & k: $3\times3$ & $32\times 32$ & $32\times 100$\\
\midrule
Conv2 & c: $64$ & k: $3\times3$ & $32\times 32$ & $32\times 100$\\
\midrule
Pool1 & k: $2\times2$ & s: $2\times2$ & $16\times 16$ & $16\times 50$\\
\midrule
\textbf{Block1}& \multicolumn{2}{c}{$\begin{bmatrix}\rm{c:}128,\rm{k:}3\times3\\\rm{c:}128,\rm{k:}3\times3\end{bmatrix}\times 1$} & $16\times 16$ & $16\times 50$\\
\midrule
Conv3 & c: $128$ & k: $3\times3$ & $16\times 16$ & $16\times 50$\\
\midrule
Pool2 & k: $2\times2$ & s: $2\times2$ & $8\times 8$ & $8\times 25$\\
\midrule
\textbf{Block2}& \multicolumn{2}{c}{$\begin{bmatrix}\rm{c:}256,\rm{k:}3\times3\\\rm{c:}256,\rm{k:}3\times3\end{bmatrix}\times 2$} & $8\times 8$ & $8\times 25$\\
\midrule
Conv4 & c: $256$ & k: $3\times3$ & $8\times 8$ & $8\times 25$\\
\midrule
\multirow{2}{*}{Pool3} & k: $2\times2$ &  & \multirow{2}{*}{$4\times 9$} & \multirow{2}{*}{$4\times 26$}\\
 & s: $2\times1$ & p: $0\times1$ &  & \\
\midrule
\textbf{Block3}& \multicolumn{2}{c}{$\begin{bmatrix}\rm{c:}512,\rm{k:}3\times3\\\rm{c:}256,\rm{k:}3\times3\end{bmatrix}\times 5$} & $4\times 9$ & $4\times 26$\\
\midrule
Conv5 & c: $512$ & k: $3\times3$ & $4\times 9$ & $4\times 26$\\
\midrule
\textbf{Block4}& \multicolumn{2}{c}{$\begin{bmatrix}\rm{c:}512,\rm{k:}3\times3\\\rm{c:}512,\rm{k:}3\times3\end{bmatrix}\times 3$} & $4\times 9$ & $4\times 26$\\
\midrule
\multirow{2}{*}{Conv6} & c: $512$ & k: $2\times2$ & \multirow{2}{*}{$2\times 10$} & \multirow{2}{*}{$2\times 27$}\\
 & s: $2\times1$ & p: $0\times1$ & & \\
\midrule
\multirow{2}{*}{Conv7} & c: $512$ & k: $2\times2$ & \multirow{2}{*}{$1\times 9$} & \multirow{2}{*}{$1\times 26$}\\
 & s: $1\times1$ & p: $0\times0$ & & \\
\bottomrule 
\end{tabular}
\end{adjustbox}
\par\end{centering}
\label{tab-ResNet}
\vspace{-0em}
\end{table}

\begin{table}[t]
\caption{Architecture of the decoder for the self-supervised learning of ResNet-29 in the Section of \textsl{Probe Evaluation}.} 
\begin{centering}
\begin{adjustbox}{width=0.44\textwidth}
\begin{tabular}[t]{c lc}
\toprule
\textbf{Layers} & \textbf{Configurations} & \textbf{Size} \\
\midrule
Input & Feature Maps & $1\times 9$ \\
\midrule
DeConv & c: $256$, k: $2\times2$, s: $1\times1$, p: $0\times0$, ReLU & $2\times 10$ \\
\midrule
Conv & c: $256$, k: $3\times3$, s: $1\times1$, p: $1\times1$, BN, ReLU & $2\times 10$ \\
\midrule
DeConv & c: $192$, k: $2\times2$, s: $2\times1$, p: $0\times0$, ReLU & $4\times 11$ \\
\midrule
Conv & c: $192$, k: $3\times3$, s: $1\times1$, p: $1\times0$, BN, ReLU & $4\times 9$ \\
\midrule
DeConv & c: $160$, k: $2\times2$, s: $2\times1$, p: $0\times0$, ReLU & $8\times 10$ \\
\midrule
Conv & c: $160$, k: $3\times3$, s: $1\times1$, p: $1\times0$, BN, ReLU & $8\times 8$ \\
\midrule
Upsample & Ratio: $\times 2$, Mode: ``nearest'' & $16\times 16$ \\
\midrule
Conv & c: $128$, k: $3\times3$, s: $1\times1$, p: $1\times1$, ReLU & $16\times 16$ \\
\midrule
Conv & c: $128$, k: $3\times3$, s: $1\times1$, p: $1\times1$, BN, ReLU & $16\times 16$ \\
\midrule
Upsample & Ratio: $\times 2$, Mode: ``nearest'' & $32\times 32$ \\
\midrule
Conv & c: $64$, k: $3\times3$, s: $1\times1$, p: $1\times1$, ReLU & $32\times 32$ \\
\midrule
Conv & c: $64$, k: $3\times3$, s: $1\times1$, p: $1\times1$, BN, ReLU & $32\times 32$ \\
\midrule
Conv & c: $3$, k: $3\times3$, s: $1\times1$, p: $1\times1$, Tanh($\cdot$) & $32\times 32$ \\
\bottomrule 
\end{tabular}
\end{adjustbox}
\par\end{centering}
\label{tab-Reg-Decoder}
\vspace{-0em}
\end{table}

\begin{table}[t]
\caption{Architecture of the decoder for the self-supervised learning of first three blocks of ResNet-29 in the Section of \textsl{Semi-Supervision Evaluation}.} 
\begin{centering}
\begin{adjustbox}{width=0.44\textwidth}
\begin{tabular}[t]{c lc}
\toprule
\textbf{Layers} & \textbf{Configurations} & \textbf{Size} \\
\midrule
Input & Feature Maps & $4\times 9$ \\
\midrule
DeConv & c: $256$, k: $2\times2$, s: $1\times1$, p: $0\times0$, ReLU & $5\times 10$ \\
\midrule
Conv & c: $256$, k: $3\times3$, s: $1\times1$, p: $1\times1$, BN, ReLU & $5\times 10$ \\
\midrule
DeConv & c: $192$, k: $2\times2$, s: $2\times1$, p: $0\times0$, ReLU & $11\times 11$ \\
\midrule
Conv & c: $192$, k: $3\times3$, s: $1\times1$, p: $0\times0$, BN, ReLU & $9\times 9$ \\
\midrule
DeConv & c: $160$, k: $2\times2$, s: $1\times1$, p: $0\times0$, ReLU & $10\times 10$ \\
\midrule
Conv & c: $160$, k: $3\times3$, s: $1\times1$, p: $0\times0$, BN, ReLU & $8\times 8$ \\
\midrule
Upsample & Ratio: $\times 2$, Mode: ``nearest'' & $16\times 16$ \\
\midrule
Conv & c: $128$, k: $3\times3$, s: $1\times1$, p: $1\times1$, ReLU & $16\times 16$ \\
\midrule
Conv & c: $128$, k: $3\times3$, s: $1\times1$, p: $1\times1$, BN, ReLU & $16\times 16$ \\
\midrule
Upsample & Ratio: $\times 2$, Mode: ``nearest'' & $32\times 32$ \\
\midrule
Conv & c: $64$, k: $3\times3$, s: $1\times1$, p: $1\times1$, ReLU & $32\times 32$ \\
\midrule
Conv & c: $64$, k: $3\times3$, s: $1\times1$, p: $1\times1$, BN, ReLU & $32\times 32$ \\
\midrule
Conv & c: $3$, k: $3\times3$, s: $1\times1$, p: $1\times1$, Tanh($\cdot$) & $32\times 32$ \\
\bottomrule 
\end{tabular}
\end{adjustbox}
\par\end{centering}
\label{tab-Reg-Decoder-Block3}
\vspace{-0em}
\end{table}

\section{Recognizer Initialization}
In the section of \textsl{Probe Evaluation}, we simply initialize the recognizer backbone using the whole pre-trained backbone parameters. This is a common setting to perform a probe evaluation to validate the representation quality. The 
architecture of the recognizer backbone (ResNet-29) and the corresponding decoder for its self-supervised training are shown in Table~\ref{tab-ResNet} and Table~\ref{tab-Reg-Decoder}, respectively.

\begin{figure}[h]
  \centering
  \includegraphics[width=0.9\columnwidth]{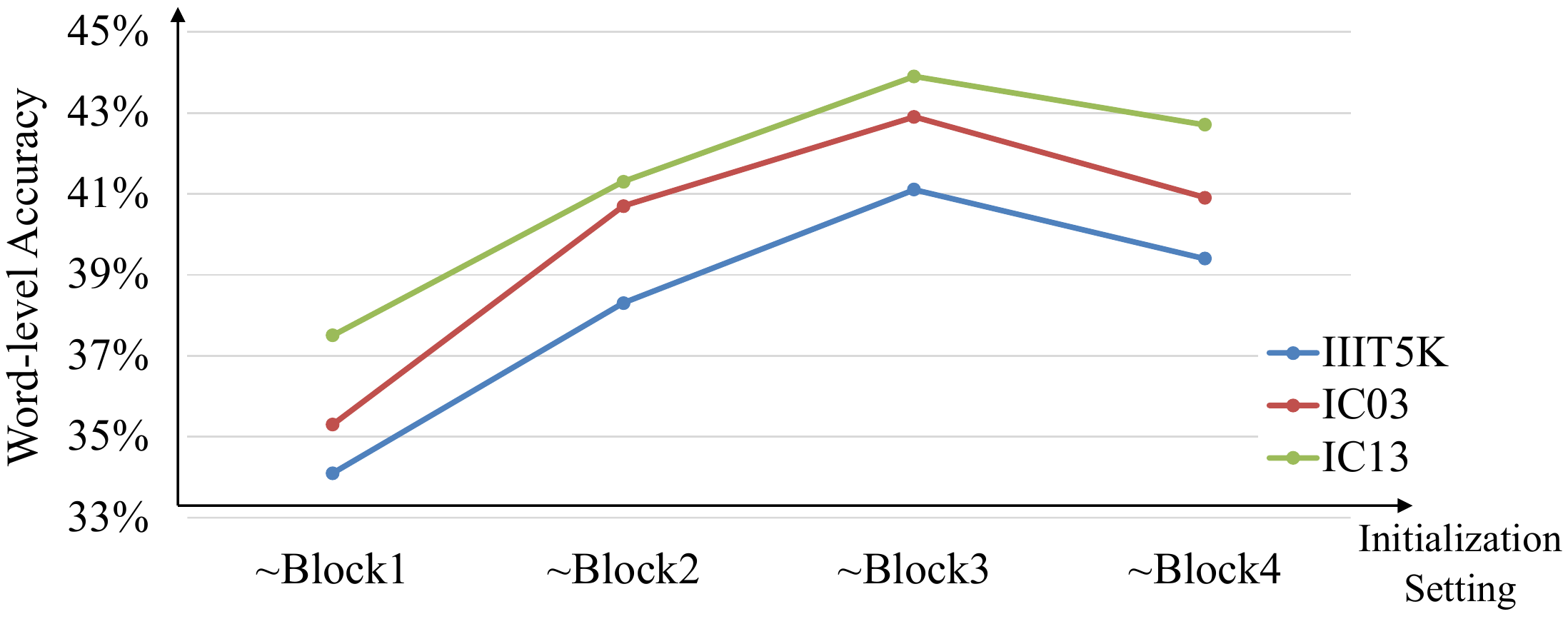}
  \caption{The recognizer achieves the best performance by using the pre-trained parameters up to a depth at ``Block3".}
  \label{pic-semi}
  \vspace{-0em}
\end{figure}

However, it is revealed by~\cite{chen2020generative} that not all the pre-trained parameters can benefit the downstream task. Therefore, for the experiment of \textsl{Semi-Supervision Evaluation}, we explore different initialization settings, \textit{i.e.}, how many layers (how deep) should be initialized by using pre-trained parameters. Specifically, we simply choose four blocks of the recognizer backbone as our four depth options. We fine-tune the recognizer using 10K labeled samples of SynthText~\cite{gupta2016synthetic}. As shown in Figure~\ref{pic-semi}, the recognizer achieves the best performance with the initialization setting at depth ``Block3". We provide the decoder for the self-supervised learning of the first three blocks, as shown in Table~\ref{tab-Reg-Decoder-Block3}.

\section{Probe/Recognizer Objectives}
After the self-supervised representation learning stage, we perform probe and semi-supervision evaluation. We set the batch size to 256 and train the recognizer for 50K iterations. The optimizer is AdaDelta~\cite{zeiler2012adadelta} with the default setting. The learning rate is set to $1.0$ and linearly decreased to $0.1$. The input word images are resized to $64 \times 200$. The experiments are conducted on the PyTorch framework~\cite{pytorch} using two NVIDIA P100 GPUs (16GB memory per GPU).

The probe/recognizer outputs 95 categories, including 52 case-sensitive letters, 10 digits, 32 punctuation symbols, and an additional ``Blank'' token for CTC decoding~\cite{graves2006connectionist} or an ``End of Sequence'' token for attention decoding~\cite{bahdanau2014neural}.

1) The CTC decoder~\cite{graves2006connectionist} transforms the feature sequence $F\in \mathbb{R} ^ {T \times C}$ to an output sequence $Y\in \mathbb{R} ^ {T \times 95}$ using a fully connected layer. For each time step, $y_t \in \mathbb{R} ^ {95}$ denotes the probability distribution over 95 categories. The objective is to minimize the negative
log-likelihood of conditional probability of ground truth $GT$:
\begin{equation}
\mathcal{L}_{CTC}=-\log p(GT|Y),
\end{equation}
where the conditional probability is defined as the sum of probabilities of all possible sequence $\pi_{i}\in\boldsymbol{\pi}$
that can be mapped onto the $GT$ (For instance, ``\texttt{-CC-{}-VVV-{}-{}-P-{}-RR}'' can be mapped onto ``\texttt{CVPR}''). It is formulated as
\begin{equation}
p(GT|Y)=\sum_{\boldsymbol{\pi}}p(\boldsymbol{\pi}|Y)=\sum_{\boldsymbol{\pi}}\prod_{t=1}^{T}Y_{t}^{\pi_{i}},
\end{equation}
where $Y_{t}^{\pi_{i}}$ denotes the predicted probability at time step $t$ with a sequence $\pi_{i}\in\boldsymbol{\pi}$. 

2) The attention decoder~\cite{bahdanau2014neural} is optimized by minimizing the negative log-likelihood of conditional probability of ground truth $GT$:
\begin{equation}
\mathcal{L}_{Att}=-\sum_{t=1}^T\log p(GT_t|y_t),
\end{equation}
where $y_t$ is the predicted probability over 95 categories at time step $t$, given by
\begin{equation}
y_{t} = \operatorname{SoftMax}({W}{s}_{t}+b).
\end{equation}
The ${s}_t$ is the hidden state at the $t$-th step, updated by
\begin{equation}
{s}_t = \operatorname{GRU}({s}_{t-1}, (y_{t-1}, {g}_{t})),
\end{equation}
where ${g}_{t}$ represents the glimpse vectors
\begin{equation}
{g}_{t} = \alpha \cdot {h}.
\end{equation}
The ${h}$ denotes the feature sequence. The ${\alpha}$ is the attention mask, expressed as
\begin{equation}
\alpha = \operatorname{SoftMax}(e),
\end{equation}
\begin{equation}
e = {w}^\mathrm{T}\operatorname{Tanh}({W}_{s}{s}_{t-1}+{W}_{h}{h}+b_{e}).
\end{equation}
Here, ${W}$, $b$, ${w}^\mathrm{T}$, ${W}_{s}$, ${W}_{h}$ and $b_{e}$ are trainable parameters. 

\begin{table*}[t]
\centering
\caption{Semi-supervised performance evaluation. We sample three orders of scales (10K, 100K, and 1M) of data from SynthText (6M). Our approach can learn high-quality representations from unlabeled data and improve the supervised baseline, especially when used with low-resource labeled data.}
\begin{adjustbox}{width=.95\textwidth}
\begin{tabular}{ c c llll lll }
\toprule
Labeled Data & Supervision & IIIT5K & SVT & IC03 & IC13 & SVT-P & CT80 & IC15 \\
\midrule
\multirow{2}{*}{10K} & Sup. & 35.0 $\pm$ 6.7 & 7.9 $\pm$ 3.4 & 37.6 $\pm$ 6.3 & 38.6 $\pm$ 6.5 & 6.8 $\pm$ 2.8 & 8.5 $\pm$ 3.2& 10.4 $\pm$ 3.5 \\
                  & Semi-Sup. & \textbf{41.1} $\pm$ 1.3 & \textbf{16.2} $\pm$ 1.4 & \textbf{42.9} $\pm$ 2.1 & \textbf{43.9} $\pm$ 1.2 & \textbf{14.2} $\pm$ 1.2 & \textbf{15.5} $\pm$ 1.7 & \textbf{17.5} $\pm$ 1.2 \\
\midrule
\multirow{2}{*}{100K} & Sup. & 72.6 $\pm$ 0.3 & 55.2 $\pm$ 1.2 & 79.4 $\pm$ 1.1 & 75.3 $\pm$ 0.8 & 45.4 $\pm$ 0.9 & 46.7 $\pm$ 1.0 & 47.6 $\pm$ 1.0\\
                  & Semi-Sup. & \textbf{73.6} $\pm$ 0.5 & \textbf{55.3} $\pm$ 1.0 & \textbf{79.9} $\pm$ 1.0 & \textbf{75.6} $\pm$ 0.5 & \textbf{45.6} $\pm$ 0.8 & \textbf{46.8} $\pm$ 1.5 & \textbf{47.9} $\pm$ 0.4 \\
\midrule
\multirow{2}{*}{1M} & Sup. & \textbf{84.1} $\pm$ 0.5 & \textbf{73.1} $\pm$ 0.2 & 88.2 $\pm$ 0.6 & 86.4 $\pm$ 1.0 & 60.5 $\pm$ 0.8 & 59.5 $\pm$ 1.5 & 58.9 $\pm$ 0.8 \\
                  & Semi-Sup. & \textbf{84.1} $\pm$ 0.6 & \textbf{73.1} $\pm$ 1.0 & \textbf{89.2} $\pm$ 1.1 & \textbf{86.5} $\pm$ 0.9 & \textbf{62.1} $\pm$ 1.1 & \textbf{63.7} $\pm$ 2.8 & \textbf{59.7} $\pm$ 0.6 \\
\midrule
\multirow{2}{*}{6M} & Sup. & 86.6 $\pm$ 0.5 & 79.6 $\pm$ 0.6 & 91.5 $\pm$ 0.7 & 89.0 $\pm$ 0.3 & \textbf{68.3} $\pm$ 0.7 & \textbf{71.9} $\pm$ 1.8 & \textbf{66.2} $\pm$ 0.6\\
                  & Semi-Sup. & \textbf{87.5} $\pm$ 0.3 & \textbf{80.6} $\pm$ 0.5 & \textbf{91.8} $\pm$ 0.7 & \textbf{89.9} $\pm$ 0.6 & \textbf{68.3} $\pm$ 1.1 & 71.4 $\pm$ 1.7 & \textbf{66.2} $\pm$ 0.4\\
\bottomrule
\end{tabular}
\end{adjustbox}
\label{tab-semi-seven}
\end{table*}

\section{Adversarial Loss}

We adopt an adversarial objective to minimize the distribution shift between the generated and real data, which is a widely used setting for image generating tasks. To study the effectiveness of the adversarial training, we conduct an ablation experiment by disabling the adversarial loss $\mathcal{L}_{adv}$. As shown in Table~\ref{tab-ablation}, the $\mathcal{L}_{adv}$ increases representation quality and makes the generated distribution closer to the real one. We believe the $\mathcal{L}_{adv}$ is necessary for visual effects, because it achieves more lifelike images.

\begin{table}[h]
\centering
\caption{Ablation study of adversarial loss. We evaluate the representation quality using an attention probe with two BiLSTMs (256 hidden units), and the distribution shift using FID~\cite{heusel2017gans} score. We average the word accuracies (\%) of IIIT5K, IC03 and IC13.}
\begin{adjustbox}{width=0.2\textwidth}
\begin{tabular}{c c c}
\toprule
$\mathcal{L}_{adv}$ & Acc. $\uparrow$ & FID $\downarrow$ \\
\midrule
$\times$ & 68.8 & 24.0 \\
\checkmark & \textbf{69.0} & \textbf{23.2} \\
\bottomrule
\end{tabular}
\end{adjustbox}
\label{tab-ablation}
\vspace{-1.em}
\end{table}

\section{Compare with AdaIN}
It is known that the AdaIN~\cite{huang2017arbitrary,karras2019style} can transfer style using global statistics (mean and standard deviation) of feature maps. We conduct a probe evaluation (following the setting of ResNet-FCN-Att.) to compare our SimAN with AdaIN. As shown in Table~\ref{tab-AdaIN}, the proposed SimAN outperforms AdaIN. This suggests the representation capability is improved by the similarity estimation, which minimizes the distance between similar patterns. 

\begin{table}[h]
\centering
\caption{Probe evaluation of AdaIN and SimAN. }
\begin{adjustbox}{width=0.45\textwidth}
\begin{tabular}{c c c c c c c}
\toprule
\multirow{2}{*}{Method} & \multicolumn{2}{c}{IIIT5K} & \multicolumn{2}{c}{IC03} & \multicolumn{2}{c}{IC13} \\ 
 \cmidrule(lr){2-3}  \cmidrule(lr){4-5}  \cmidrule(lr){6-7}
 & Acc. & E.D. 1 & Acc. & E.D. 1 & Acc. & E.D. 1 \\ 
\midrule
AdaIN & 9.7 & 21.9 & 9.1 & 18.7 & 11.1 & 24.6 \\
SimAN & \textbf{22.2} & \textbf{39.7} & \textbf{22.3} & \textbf{38.6} & \textbf{24.1} & \textbf{43.6} \\
\bottomrule
\end{tabular}
\end{adjustbox}
\label{tab-AdaIN}
\vspace{-0.5em}
\end{table}

\section{Semi-Supervision Evaluation}
We provide experimental results of five runs on seven popular benchmarks in Table~\ref{tab-semi-seven}.

\section{Network Architecture}

We present the encoder and decoder used in the Section of \textsl{Generative Visual Task} in Table~\ref{tab-Encoder} and~\ref{tab-Decoder}, respectively. These are popular architectures and are widely used~\cite{huang2017arbitrary,johnson2016perceptual}.

We present the discriminator in Table~\ref{tab-discriminator}.

\begin{table}[t]
\caption{Architecture of the encoder in the Section of \textsl{Generative Visual Task}. } 
\begin{centering}
\begin{adjustbox}{width=0.48\textwidth}
\begin{tabular}[t]{c lll c}
\toprule
\textbf{Layers} & \multicolumn{3}{c}{\textbf{Configurations}} & \textbf{Size} \\
\midrule
Input & \multicolumn{3}{c}{RGB image} & $3 \times 64\times 64$ \\
\midrule
Conv1 & c: $3$ & k: $1$ & & $3 \times 64\times 64$ \\
\midrule
Conv2 & c: $64$ & k: $3$ & Reflection Pad: $1$, ReLU& $64 \times 64\times 64$ \\
\midrule
Conv3 & c: $64$ & k: $3$ & Reflection Pad: $1$, ReLU& $64 \times 64\times 64$ \\
\midrule
MaxPool & k: $2$ & s: $2$ & & $64 \times 32\times 32$ \\
\midrule
Conv4 & c: $128$ & k: $3$ & Reflection Pad: $1$, ReLU& $128 \times 32\times 32$ \\
\midrule
Conv5 & c: $128$ & k: $3$ & Reflection Pad: $1$, ReLU& $128 \times 32\times 32$ \\
\midrule
MaxPool & k: $2$ & s: $2$ & & $128 \times 16\times 16$ \\
\midrule
Conv6 & c: $256$ & k: $3$ & Reflection Pad: $1$, ReLU& $256 \times 16\times 16$ \\
\midrule
Conv7 & c: $256$ & k: $3$ & Reflection Pad: $1$, ReLU& $256 \times 16\times 16$ \\
\midrule
Conv8 & c: $256$ & k: $3$ & Reflection Pad: $1$, ReLU& $256 \times 16\times 16$ \\
\midrule
Conv9 & c: $256$ & k: $3$ & Reflection Pad: $1$, ReLU& $256 \times 16\times 16$ \\
\midrule
MaxPool & k: $2$ & s: $2$ & & $256 \times 8\times 8$ \\
\midrule
Conv10 & c: $512$ & k: $3$ & Reflection Pad: $1$, ReLU& $512 \times 8\times 8$ \\
\bottomrule 
\end{tabular}
\end{adjustbox}
\par\end{centering}
\label{tab-Encoder}
\vspace{0.5em}
\end{table}

\section{Data Synthesis}
Following the pipeline of SynthText~\cite{gupta2016synthetic}, we simply render a text on a clean canvas. The fonts are publicly available\footnote{\url{https://fonts.google.com/}}. We follow the strict setting proposed by Long \textit{et al.}~\cite{long2020unrealtext} to include punctuation symbols, digits, upper-case and lower-case characters for evaluation. We also use the same recognizer trained on our 1M synthetic data.

We perform random blurring on the synthetic data to meet the low-quality practice of scene text images. The pseudo-code is shown as below for reference.

\begin{lstlisting}[language=Python]
import albumentations as A
A.OneOf([
    A.ImageCompression(40, 80),
    A.Blur(blur_limit=[5, 11]),
    A.GaussianBlur(blur_limit=(5, 11)),
    A.MedianBlur(blur_limit=[5, 11]),
    A.MotionBlur(blur_limit=[5, 11])
])
\end{lstlisting}

\begin{table}[h]
\caption{Architecture of the decoder in the Section of \textsl{Generative Visual Task}. } 
\begin{centering}
\begin{adjustbox}{width=0.48\textwidth}
\begin{tabular}[t]{c lll c}
\toprule
\textbf{Layers} & \multicolumn{3}{c}{\textbf{Configurations}} & \textbf{Size} \\
\midrule
Input & \multicolumn{3}{c}{Feature Map} & $512 \times 8\times 8$ \\
\midrule
Conv1 & c: $256$ & k: $3$ & Reflection Pad: $1$, ReLU & $256 \times 8\times 8$ \\
\midrule
Upsample & \multicolumn{3}{c}{Ratio: $\times 2$, Mode: ``nearest''} & $256 \times 16\times 16$ \\
\midrule
Conv2 & c: $256$ & k: $3$ & Reflection Pad: $1$, ReLU & $256 \times 16\times 16$ \\
\midrule
Conv3 & c: $256$ & k: $3$ & Reflection Pad: $1$, ReLU & $256 \times 16\times 16$ \\
\midrule
Conv4 & c: $256$ & k: $3$ & Reflection Pad: $1$, ReLU & $256 \times 16\times 16$ \\
\midrule
Conv5 & c: $128$ & k: $3$ & Reflection Pad: $1$, ReLU & $128 \times 16\times 16$ \\
\midrule
Upsample & \multicolumn{3}{c}{Ratio: $\times 2$, Mode: ``nearest''} & $128 \times 32\times 32$ \\
\midrule
Conv6 & c: $128$ & k: $3$ & Reflection Pad: $1$, ReLU & $128 \times 32\times 32$ \\
\midrule
Conv7 & c: $64$ & k: $3$ & Reflection Pad: $1$, ReLU & $64 \times 32\times 32$ \\
\midrule
Upsample & \multicolumn{3}{c}{Ratio: $\times 2$, Mode: ``nearest''} & $64 \times 64\times 64$ \\
\midrule
Conv8 & c: $64$ & k: $3$ & Reflection Pad: $1$, ReLU & $64 \times 64\times 64$ \\
\midrule
Conv9 & c: $3$ & k: $3$ & Reflection Pad: $1$, ReLU & $3 \times 64\times 64$ \\
\midrule
Tanh & \multicolumn{3}{c}{-} & $3 \times 64\times 64$ \\
\bottomrule 
\end{tabular}
\end{adjustbox}
\par\end{centering}
\label{tab-Decoder}
\vspace{0.5em}
\end{table}

\begin{table}[h]
\centering
\caption{Architecture of the discriminator.}
\begin{centering}
\begin{adjustbox}{width=0.36\textwidth}
\begin{tabular}[t]{c lllll}
\toprule
\textbf{Layers} & \multicolumn{5}{c}{\textbf{Configurations}} \\
\midrule
Conv1 & c: $64$ & k: $4$ & s: $2$ & p: $1$ & PReLU \\
\midrule
Conv2 & c: $128$ & k: $4$ & s: $2$ & p: $1$ & PReLU \\
\midrule
Conv3 & c: $256$ & k: $4$ & s: $2$ & p: $1$ & PReLU \\
\midrule
Conv4 & c: $512$ & k: $4$ & s: $1$ & p: $1$ & PReLU \\
\midrule
Conv5 & c: $1$ & k: $4$ & s: $1$ & p: $1$ & \\
\bottomrule 
\end{tabular}
\end{adjustbox}
\par\end{centering}
\label{tab-discriminator}
\end{table}

\begin{figure}[!]
  \centering
  \includegraphics[width=1.01\linewidth]{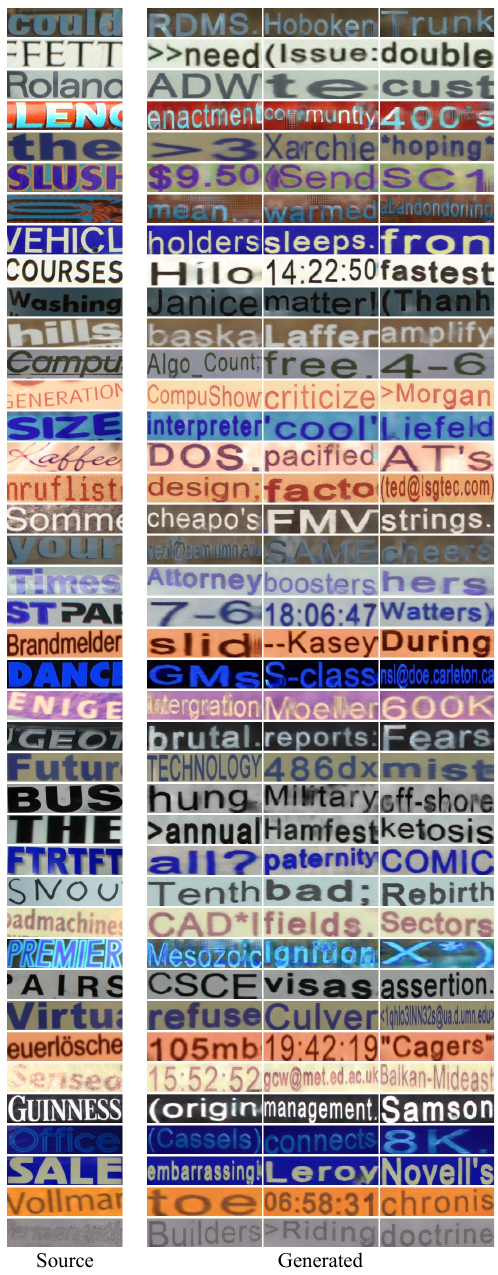}
  \caption{Arbitrary-length text editing. All the images are resized to $(64, 256)$.}
  \label{pic-edit-supp}
  \vspace{-0em}
\end{figure}

\section{Arbitrary-Length Text Editing}
We follow the same setting as EditText~\cite{wu2019editing} to generate a content image. We use a standard font style ``Arial.ttf'' to put the target text string on a clean canvas as content input. The target text is randomly selected from the corpus of SynthText~\cite{gupta2016synthetic}. Thus, the length of the source text string and the target one can be significantly different, which simulates practice challenges. We present the edited results in Figure~\ref{pic-edit-supp}.

\section{Font Interpolation}
We formulate the process of font interpolation as mathematical equations. First, we extract the content representations of the source image and target image as $Q$ and $K$, respectively. Besides, we obtain their style representations $\mu^{\operatorname{source}}$, $\sigma^{\operatorname{source}}$, $\mu^{\operatorname{target}}$ and $\sigma^{\operatorname{target}}$.

\subsection{Color Interpolation}
First, we rearrange the target style representations according to the source content representation $Q$:
\begin{equation}
\small
\begin{split}
&\mu^{\operatorname{rearrange}}=\mu^{\operatorname{target}} \operatorname{Softmax}\left(\frac{K^{\mathrm{T}}Q}{\sqrt{d_{k}}}\right),\\
&\sigma^{\operatorname{rearrange}}=\sigma^{\operatorname{target}} \operatorname{Softmax}\left(\frac{K^{\mathrm{T}}Q}{\sqrt{d_{k}}}\right).
\end{split}
\end{equation}

Then we perform interpolation on the source and rearranged style representations:
\begin{equation}
\small
\begin{split}
&\mu^{\prime} = (1 - \alpha) \mu^{\operatorname{source}} + \alpha \mu^{\operatorname{rearrange}},\\
&\sigma^{\prime} = (1 - \alpha) \sigma^{\operatorname{source}} + \alpha \sigma^{\operatorname{rearrange}}, \alpha \in [0, 1].
\end{split}
\end{equation}

Finally, we decode the feature maps to obtain an image:
\begin{equation}
\small
Q_{c, i, j}^{\prime} = Q_{c, i, j}\sigma_{c, i, j}^{\prime} + \mu_{c, i, j}^{\prime},
\end{equation}
\begin{equation}
\small
I_{rec} = \operatorname{Decoder}(Q^{\prime}).
\end{equation}

\subsection{Glyph Interpolation}
The glyph interpolation requires a same character/string on the source and target image. First, we normalize the target glyph image using the source style:
\begin{equation}
\small
\begin{split}
&\mu^{\operatorname{rearrange}}=\mu^{\operatorname{source}} \operatorname{Softmax}\left(\frac{Q^{\mathrm{T}}K}{\sqrt{d_{k}}}\right),\\
&\sigma^{\operatorname{rearrange}}=\sigma^{\operatorname{source}} \operatorname{Softmax}\left(\frac{Q^{\mathrm{T}}K}{\sqrt{d_{k}}}\right).
\end{split}
\end{equation}

The target glyph can be presented as:
\begin{equation}
\small
K_{c, i, j}^{\operatorname{rearrange}} = K_{c, i, j}\sigma_{c, i, j}^{\operatorname{rearrange}} + \mu_{c, i, j}^{\operatorname{rearrange}}.
\end{equation}
This ensures the difference between the source and target representation is only the glyph.

Then we perform interpolation on the source and rearranged glyph representation:
\begin{equation}
\small
Q_{c, i, j}^{\prime} = (1 - \alpha) Q_{c, i, j} + \alpha K_{c, i, j}^{\operatorname{rearrange}}, \alpha \in [0, 1].
\end{equation}

Finally, we obtain an image:
\begin{equation}
\small
I_{rec} = \operatorname{Decoder}(Q^{\prime}).
\end{equation}

\end{document}